\documentclass[10pt,twocolumn,letterpaper]{article}

\usepackage[pagenumbers]{iccv} 

\definecolor{iccvblue}{rgb}{0.21,0.49,0.74}
\usepackage[pagebackref,breaklinks,colorlinks,allcolors=iccvblue]{hyperref}

\usepackage{setspace}
\usepackage{multirow}
\usepackage{colortbl}
\usepackage[dvipsnames]{xcolor}
\usepackage{graphicx}
\usepackage{bm}
\usepackage{makecell}
\usepackage{booktabs}
\usepackage{tabularx}
\usepackage{tcolorbox}
\usepackage{color}
\usepackage{tcolorbox}
\usepackage{hyperref}
\usepackage{fancyvrb}
\usepackage{tikz}
\usepackage{pifont}
\definecolor{lightgray}{gray}{0.9}
\definecolor{lightblue}{rgb}{0.88,0.95,1}
\definecolor{lightgreen}{rgb}{0.88,1,0.88}
\definecolor{darkblue}{rgb}{0.2,0.4,0.8}
\definecolor{darkred}{rgb}{0.8,0.2,0.2}
\definecolor{darkgreen}{rgb}{0.0,0.5,0.0}

\newcommand{\xmark}{\ding{55}}%

\newcommand{\drawImageWithBoxes}[4]{
    \begin{center}
    \begin{tikzpicture}
        \node[anchor=south west, inner sep=0] (img) at (0,0) {\includegraphics[width=\linewidth]{#1}};

        \begin{scope}[x={(img.south east)},y={(img.north west)}]
        
            #4
            
        \end{scope}
    \end{tikzpicture}
    \end{center}
}

\newcommand{\drawBoundingBox}[7]{
    \draw[#7, ultra thick] (#1/#5,1-#2/#6) rectangle ++(#3/#5,-#4/#6);
}

\def\paperID{2588} 
\def\confName{ICCV}
\def\confYear{2025}

\title{Retrieval-Based Interleaved Visual Chain-of-Thought \\ in Real-World Driving Scenarios}

\author{
  Charles Corbière$^*$ \quad Simon Roburin$^*$ \quad Syrielle Montariol\thanks{Equal contribution, with order determined alphabetically.} \\ Antoine Bosselut \quad Alexandre Alahi \\[5pt]
  {\normalsize EPFL, Switzerland} \\
}

\begin{document}

\maketitle
\begin{abstract}

While chain-of-thought (CoT) prompting improves reasoning in large language models, its effectiveness in vision-language models (VLMs) remains limited due to over-reliance on textual cues and memorized knowledge. To investigate the visual reasoning capabilities of VLMs in complex real-world scenarios, we introduce \textsc{DrivingVQA}, a visual question answering dataset derived from driving theory exams, which contains 3,931 multiple-choice problems with expert-written explanations and grounded entities relevant to the reasoning process. Leveraging this dataset, we propose \textit{RIV-CoT}, a Retrieval-Based Interleaved Visual Chain-of-Thought method that enables VLMs to reason using visual crops corresponding to these relevant entities. Our experiments demonstrate that RIV-CoT improves answer accuracy by 3.1\% and reasoning accuracy by 4.6\% over vanilla CoT prompting. Furthermore, we demonstrate that our method effectively scales to the larger A-OKVQA reasoning dataset by leveraging automatically generated pseudo-labels, outperforming CoT prompting. Code and dataset are available at \href{https://vita-epfl.github.io/DrivingVQA}{https://vita-epfl.github.io/DrivingVQA}.

\end{abstract}    
\section{Introduction}
\label{sec:intro}

 \begin{figure}[ht]
  \centering
    \includegraphics[width=\linewidth]{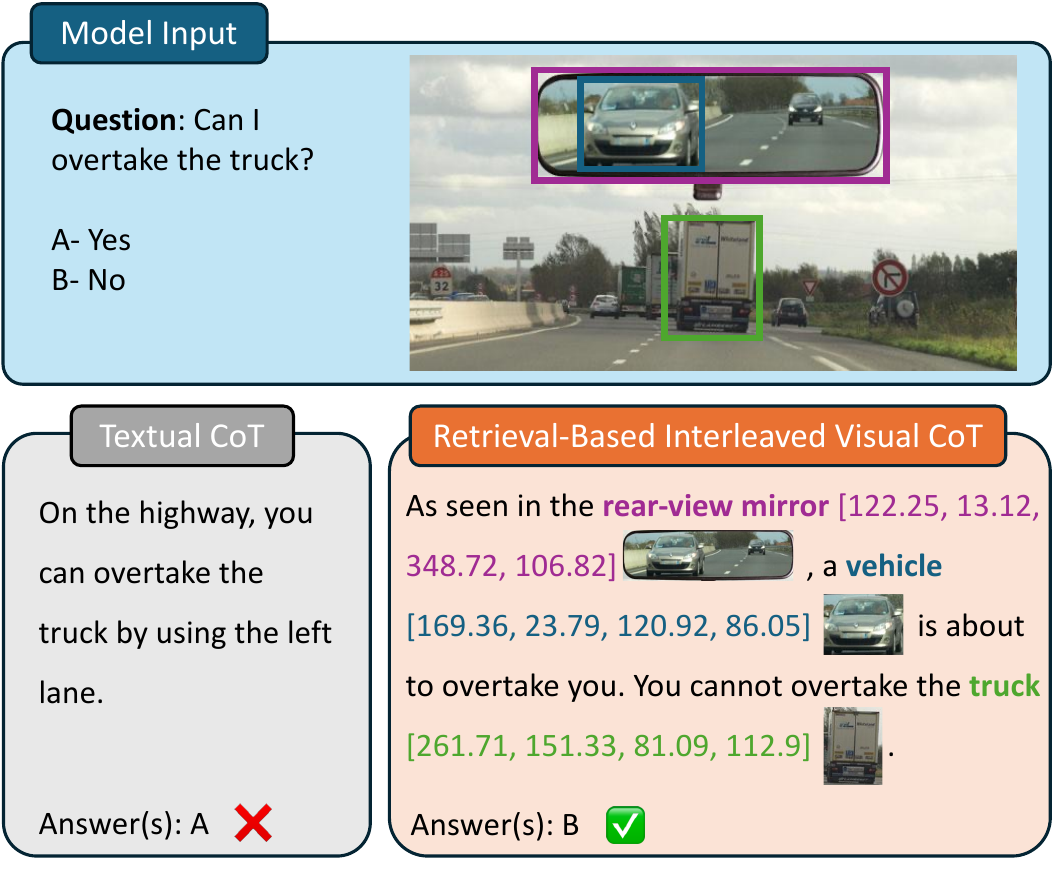}
    \caption{\textbf{Illustration of retrieval-based interleaved visual chain-of-thought in \textsc{DrivingVQA}}. Successfully answering the question requires detecting relevant entities (e.g., the truck, the car in the rear-view mirror), recognizing their attributes (e.g., the car signaling to overtake), and reasoning spatially to determine whether overtaking is safe. The interleaved explanation provides step-by-step reasoning aligned with visual content.}
    \label{fig:intro}
\end{figure}

Chain-of-thought (CoT)~\citep{CoT2022} is a prompting strategy that aims at enhancing the reasoning capabilities of large language models (LLMs)~\citep{openai2024gpt4o,touvron2023llama,jiang2023mistral7b} and, more recently, vision-language models (VLMs)~\citep{Qwen2VL,liu2024visual,alayrac2022flamingo,lin2024vila}. While well-suited for mathematical and logical reasoning, CoT has limited effectiveness when it comes to visual, spatial, and multi-step reasoning~\citep{sprague2024cot,zhang2024multimodal,shao2024visual}.

A critical challenge is VLMs’ tendency to excessively rely on textual inputs and memorized knowledge rather than visual inputs, which can lead to hallucinations~\citep{Guan_2024_CVPR, liu2024survey, jiang2024hallucination, chandhok2024response}. Recent methods attempt to enforce explicit grounding by generating textual image descriptions~\citep{zhangmultimodal, ozdemir2024enhancing, li2024enhancing}, leveraging scene graphs~\citep{mitra2024compositional}, or using bounding box coordinates~\citep{lei2024scaffolding, liu2025spatialcot, chen2023shikra}. 
However, these methods have limited expressiveness; textual descriptions or sets of bounding box coordinates are poorly leveraged by VLMs and fail to capture the complexity of a visual scene. 

In this work, we investigate the ability of VLMs to handle tasks that involve complex visual reasoning by drawing inspiration from cognitive research on human reasoning~\citep{piaget1978piaget}. Humans naturally engage in visual reasoning to navigate and interact with their surroundings, leveraging core cognitive abilities such as perception, spatial reasoning, and decision-making. Inspired by these mechanisms, we implement \emph{visual chain-of-thought} as a structured reasoning process that involves the following steps: detecting relevant entities in the scene, identifying their attributes, modeling spatial relationships, and reasoning to determine appropriate actions.

Existing reasoning-oriented visual question answering (VQA) datasets are insufficient for evaluating these capabilities. Some rely on synthetic or schematic visuals~\citep{chia2024puzzlevqa}, lack grounding annotations~\citep{lu2022learn,liu2025mmbench}, or present oversimplified scenarios with a single region of interest~\citep{shao2024visual}. Additionally, most explanations are generated from a set of predefined templates or by using an LLM~\citep{chia2024puzzlevqa, shao2024visual, marcu2024lingoqa}. These explanations can be repetitive, error-prone, and biased, limiting their effectiveness as training signals.

To overcome these limitations, we introduce \textsc{DrivingVQA}, a visual reasoning dataset derived from publicly available French driving theory exams. It contains 3,931 samples, each featuring one or two visual questions with multiple answer choices, alongside expert-written explanations and human annotations of relevant entities with bounding box coordinates. \textsc{DrivingVQA} offers a comprehensive framework to guide and assess VLMs’ capabilities in complex real-world scenarios that involve multi-object and spatial reasoning.

Using \textsc{DrivingVQA}, we conduct a structured analysis to investigate the impact of various prompting and training strategies on VLM’s reasoning capabilities, and propose \textit{RIV-CoT}: \textit{Retrieval-based Interleaved Visual Chain-of-Thought prompting} (see \cref{fig:intro}). While state-of-the-art VLMs struggle in zero-shot settings, we show that training strategies incorporating relevant entities information, in particular through visual crops interleaved with textual reasoning, significantly improve the correctness of reasoning and answer compared to direct answer generation and vanilla CoT prompting.

Our main contributions are as follows:
\begin{itemize}
\item We introduce \textsc{DrivingVQA}, a real-world VQA dataset designed to benchmark and enhance the visual reasoning capabilities of VLMs (\cref{sec:dataset}).
\item We propose \textit{RIV-CoT}, a retrieval-based interleaved visual chain-of-thought method that allows VLMs to reason using visual crops retrieved from the input image (\cref{sec:method});
\item We show that incorporating relevant entities information improves answer accuracy by up to 3.1\% and reasoning accuracy by up to 4.6\% compared to vanilla chain-of-thought prompting (\cref{subsec:learning_to_generate}).
\item We demonstrate that our approach scales effectively to a larger dataset that does not include relevant entities annotation, A-OKVQA~\cite{aokvqa2022}, by using automatically-generated pseudo-labels (\cref{subsec:scaling}).
\end{itemize}
\section{Related Work}
\label{sec:related_work}

\subsection{Visual Chain-of-Thought in VLMs}

A major challenge in visual CoT for VLMs is their capacity to leverage visual inputs during the reasoning process. 
Significant research efforts have been made to enhance visual CoT capabilities. One of the main strategies is to generate a textual description of the scene before answering the question~\citep{zhangmultimodal, ozdemir2024enhancing, li2024enhancing, qiao2025prism}. The description may take the form of captions, a series of visual question-answer pairs that extract additional image details~\citep{zheng2023ddCoT}, or sparser representations of the input image such as scene graphs~\citep{mitra2024compositional}. 
Similarly, visual programming~\citep{gupta2023visual, suris2023vipergpt, li2025chain} adopts a neuro-symbolic approach that leverages off-the-shelf models to extract information from images and convert it into text for subsequent reasoning. 

Another research direction focuses on explicit grounding within the input image. This includes training VLMs to generate CoT along with coordinates of relevant image regions~\citep{liu2025spatialcot, chen2023shikra}, or incorporating visual prompts directly into the image, to be used by the VLM when reasoning. These visual prompts can take the form of a grid~\citep{lei2024scaffolding} or a set of visual markers~\citep{yan2024list} that indicate specific areas of the image. More closely related to our work, CogCOM \cite{qi2024cogcom} modifies the input image before using the transformed version to answer a question, notably by zooming on a specific image region. Concurrently, Visual CoT \cite{shao2024visual} designs a visual sampler that selects a sub-region of the input image to answer a question. 
In contrast, our method teaches VLMs to perform interleaved CoT with image patches retrieved from the input image, ultimately enabling more integrated visual reasoning.


\subsection{Visual Reasoning Datasets}

Since the seminal work of \citet{VQA2015}, visual question answering (VQA) datasets \citep{GoyalKSBP16,singh2019towards,okvqa2019,hudson2019gqa} have flourished over the past decade. 
With the advent of VLMs, recent benchmarks~\citep{lu2022learn,chia2024puzzlevqa,shao2024visual,yue2023mmmu} are designed to evaluate deeper reasoning capabilities, particularly across multiple disciplines with domain-specific knowledge. Among datasets that provide explanations in addition to the question-answer (QA) pair, many comprise synthetic images with human-annotated explanations or real-world images paired with synthetically generated explanations. For instance, ScienceQA~\citep{lu2022learn} offers human-annotated explanations sourced from elementary and high school science exams, but relies on figures, puzzles, or synthetic visuals. Similarly, PuzzleQA \cite{chia2024puzzlevqa} is generated from a finite set of predefined templates and comprises synthetic visual puzzles, questions as well as explanations. Causal-VidQA~\citep{li2022from} presents video-based questions accompanied by human-written explanation that require causal reasoning to be answered. However, it lacks the extensive entity annotations needed to ground reasoning in visual elements. GQA-CoT~\citep{shao2024visual} contains real-world images paired with their corresponding relevant entities but relies on synthetic explanations and restricts its annotations to a single entity per image. 

In the field of autonomous driving, the need for explainable driving behavior in control planning~\citep{xu2024drivegpt4,wang2024omnidriveholisticllmagentframework} and end-to-end driving~\citep{hwang2024emma,wang2023driveanywheregeneralizableendtoend} has led to the creation of many VQA datasets~\citep{kim2018textual,qian2024nuscenesqa,marcu2024lingoqa} building upon existing autonomous driving datasets~\citep{qian2024nuscenesqa,sima2023drivelm}. NuScenesQA~\citep{qian2024nuscenesqa} provides simple, single-word language responses per question which hinders the possibility of complex reasoning. The recent LingoQA benchmark~\citep{marcu2024lingoqa} comes closest to our work, featuring QA pairs annotated on an in-house dataset with textual descriptions of driver actions, justifications, and road observations. However, its questions and answers are synthetically generated by GPT-4~\citep{openai2024gpt4technicalreport} and do not include any bounding boxes or relevant entities to visually ground explanations. In contrast, \textsc{DrivingVQA} features real-world samples enriched with human-expert annotations of relevant entities to answer the question and explanations (see \cref{appx_sec:benchmarks_comparison} for a detailed dataset comparison). 

\section{\textsc{DrivingVQA} Dataset}
\label{sec:dataset}

\begin{figure}[t]
    \centering
    \includegraphics[width=\linewidth]{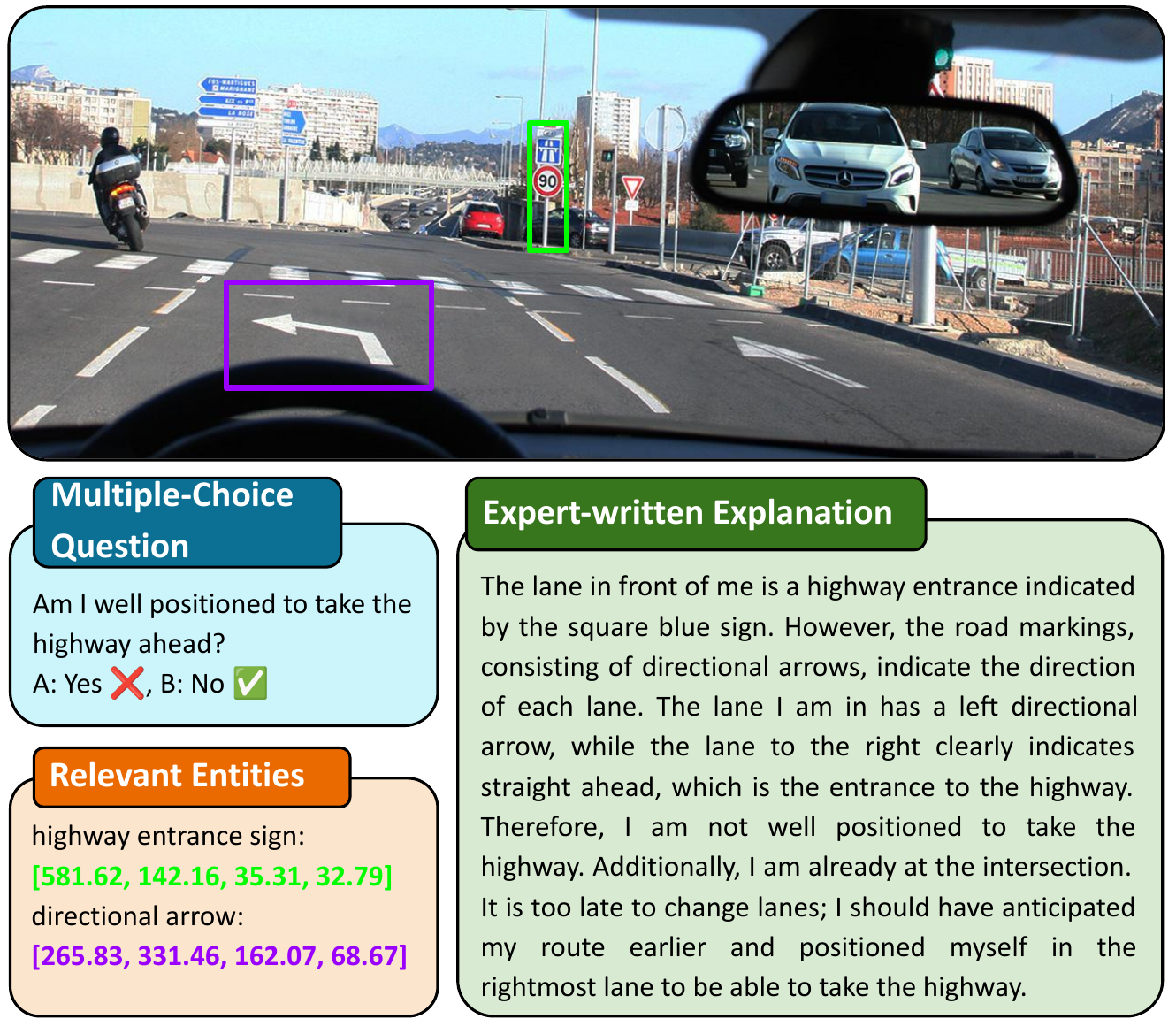}
    \caption{\textbf{\textsc{DrivingVQA} example} with a multiple-choice question, a set of relevant entities with their coordinates, and an expert-written explanation describing the situation step by step.}
    \label{fig:dataset-creation}
\end{figure}

\begin{figure}[t]
    \centering
    \begin{subfigure}[b]{0.49\linewidth}
        \centering
        \includegraphics[width=\linewidth]{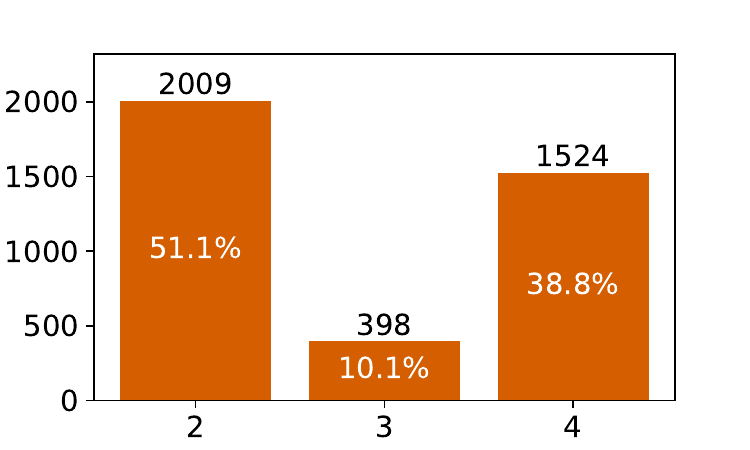}
        \caption{Number of possible answers.}
        \label{fig:subfig1}
    \end{subfigure}
    \hfill
    \begin{subfigure}[b]{0.49\linewidth}
        \centering
        \includegraphics[width=\linewidth]{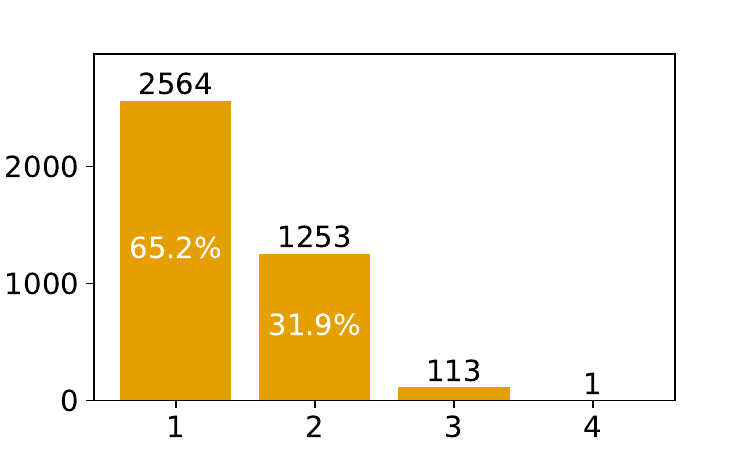}
        \caption{Number of correct answers.}
        \label{fig:subfig2}
    \end{subfigure}
    \hfill
    \begin{subfigure}[b]{0.49\linewidth}
        \centering
        \includegraphics[width=\linewidth]{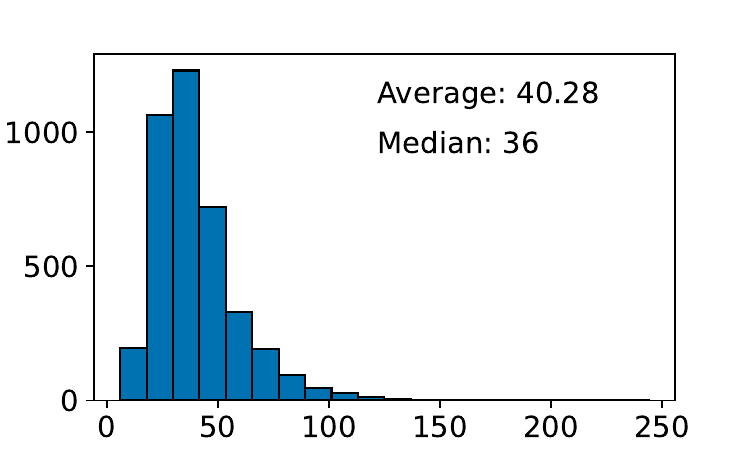}
        \caption{Number of words in expl.}
        \label{fig:subfig3}
    \end{subfigure}
    \begin{subfigure}[b]{0.49\linewidth}
        \centering
        \includegraphics[width=\linewidth]{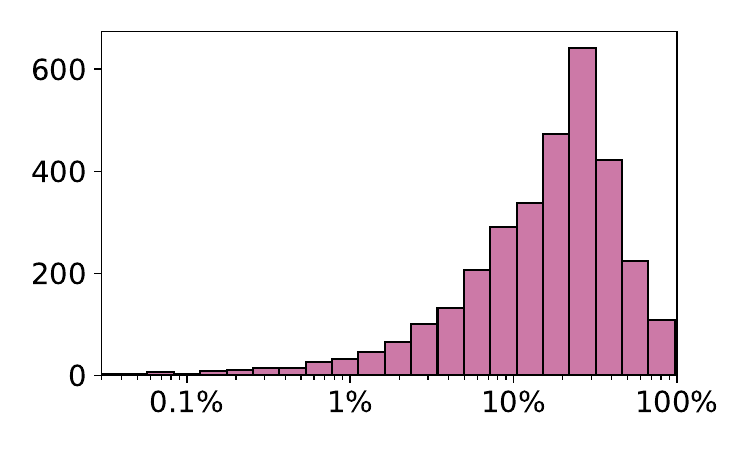}
        \caption{Image coverage by entities.}
    \end{subfigure}
    \begin{subfigure}[b]{0.49\linewidth}
        \centering
        \includegraphics[width=\linewidth]{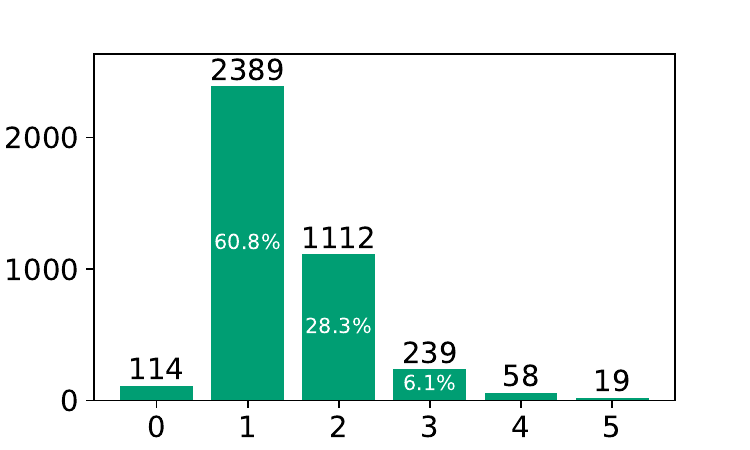}
        \caption{Number of relevant entities.}
        \label{fig:subfig4}
    \end{subfigure}
    \hfill
    \begin{subfigure}[b]{0.49\linewidth}
        \centering
\includegraphics[width=\linewidth]{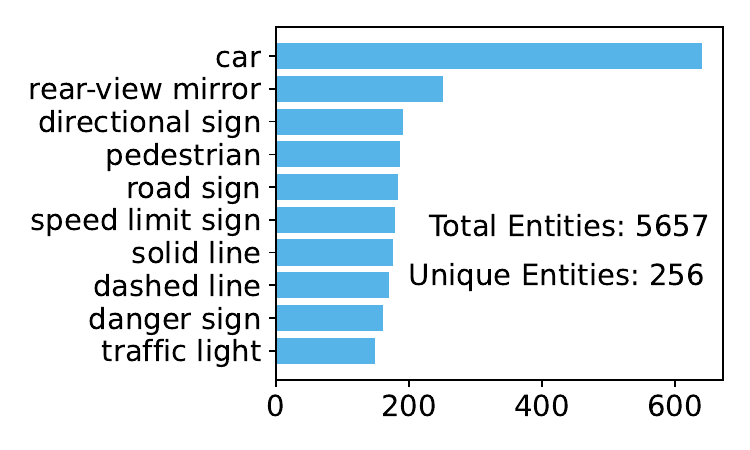}
        \caption{Top-10 most frequent entities.}
        \label{fig:subfig5}
    \end{subfigure}
    \caption{\textbf{\textsc{DrivingVQA} dataset statistics.}
    }
    \label{fig:data_stats}
\end{figure}
        
We construct \textsc{DrivingVQA} through a two-step process: (1) collecting and filtering challenging real-world driving scenarios from driving theory tests (see \cref{subsec:data_collection}), and (2) annotating relevant entities in each sample (see \cref{subsec:relevant_entities}).

\begin{figure*}[t]
    \centering
    \includegraphics[width=\linewidth,height=9cm,keepaspectratio]{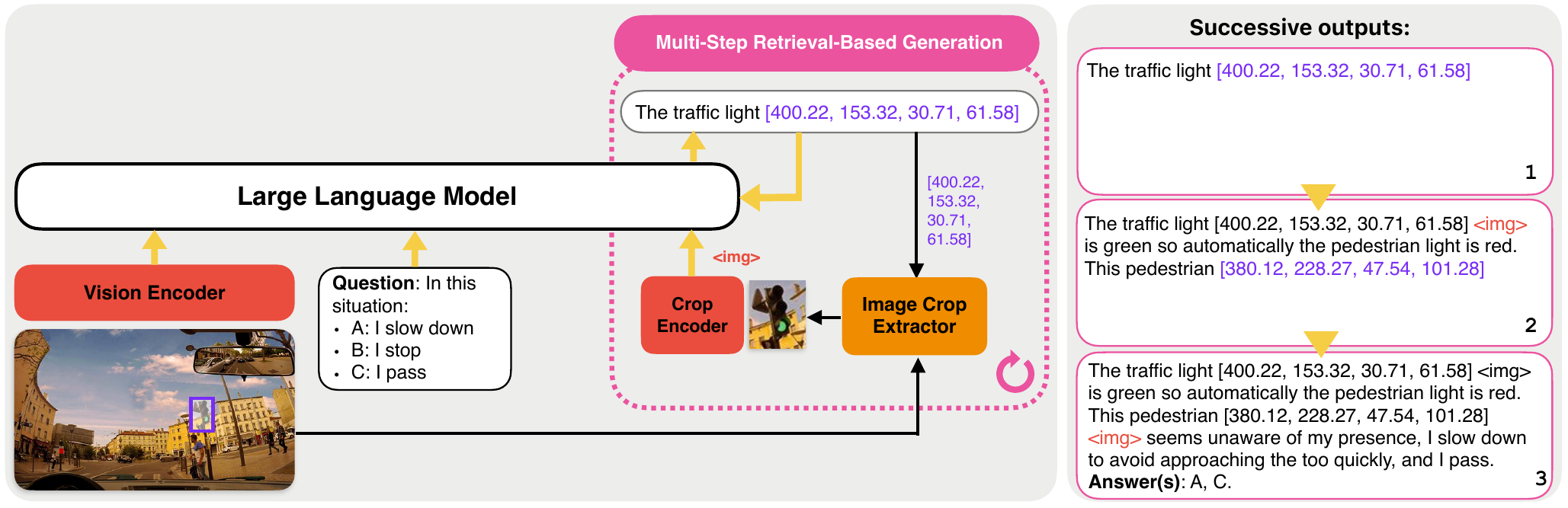}
    \caption{\textbf{Illustration of multi-step retrieval-based generation.} During inference, starting with a tokenized question and an image tokenized by the adapter on the output of the Vision Encoder, the Large Language Model generates output until it predicts a bounding box. At this point, the generation process pauses to extract the corresponding image crop based on the predicted coordinates. The image crop is encoded and adapted into an image crop token, which is then added back into the model’s context along with the question, image tokens and previously generated outputs. This iterative process continues until the model produces its final answer.}
    \label{fig:model_figure}
\end{figure*}

\subsection{Data Collection}
\label{subsec:data_collection}


In many countries, obtaining a driver’s license requires passing two examinations: a theoretical test and a practical driving test. 
In France, the theoretical exam consists of 40 multiple-choice questions (MCQs) with images. Candidates must possess accurate knowledge of traffic laws, road signs, and safe driving practices and must carefully analyze the image provided to answer the question. Given this emphasis on visual analysis and the use of real-world images, we collect French driving theory material.

\vspace{0.2cm}
\noindent
\textbf{Collection.}
To help candidates prepare for this exam, several online platforms offer practice driving theory problems. These resources are often freely accessible and do not require registration. We curate MCQs from such platforms. 
Each collected sample includes questions with two to four possible answers, where multiple answers may be correct. Some questions include two sub-questions, each with two possible answers. We standardize the format to include the image of the driving scene, the question's text, the list of possible answers, the correct answer(s), and an expert-annotated explanation. An example of a driving theory problem is shown in \cref{fig:dataset-creation}.

\vspace{0.2cm}
\noindent
\textbf{Filtering.} Not all collected questions require visual reasoning about driving scenes, and/or some images may serve only illustrative purposes. We use GPT-4o to filter out samples where questions could be answered correctly without using the image, and perform a manual review of excluded images to address potential misclassifications. In a validation exercise with 60 randomly selected samples, three expert annotators achieved an inter-annotator agreement score of 0.95 (Krippendorff's alpha, Fleiss' kappa, and Cohen's kappa).

\vspace{0.2cm}
\noindent
\textbf{Translation.} We translate all questions, answer options, and explanations from French to English using GPT-4o-mini, then manually review and refine them for consistency and fidelity to the original intent. \\

After filtering, the final \textsc{DrivingVQA} dataset contains 3,931 samples, divided into a training set (80\%) and a test set (20\%, 789 samples). \cref{fig:data_stats} provides an overview of the dataset statistics, including the distribution of the number of possible answers, correct answers, and explanation length in terms of word count.

\subsection{Relevant Entities}
\label{subsec:relevant_entities}
Here, we recall our formalization of visual CoT: detecting relevant entities in the scene, identifying their attributes, modeling spatial relationships, and reasoning to determine appropriate actions. To support this, 
We augment \textsc{DrivingVQA} with annotations of relevant entities to answer the question, specifying their names and locations in the image.

\vspace{0.2cm}
\noindent
\textbf{Automated extraction.} To reduce the burden of manual annotation, we develop a pipeline that identifies possible relevant entities and their location in the image. This process begins by leveraging human explanations to extract an initial list of domain-specific key entities found in our dataset. For each sample, we identify the entities from this list that are visible in the images and referenced in the question, possible answers, and explanation using GPT4o-mini. Then, we use GroundingDINO~\cite{liu2023grounding} to localize these entities in the image, obtaining (entity label, bounding box coordinates) pairs. Finally, we apply heuristics to refine these outputs, such as grouping similar labels under unified entity names. Details of this pipeline are provided in \cref{appx_subsec:auto_extract_pipeline}.

\vspace{0.2cm}
\noindent
\textbf{Manual annotation.} Human experts refine the pseudo-annotated data by removing irrelevant entities, correcting inaccurate labels and coordinates, and adding missing entities. In total, 5,657 entities (spanning 256 unique labels) were annotated with precise bounding boxes, averaging 1.4 entities per image. \cref{fig:data_stats} shows the distribution of entities, their image coverage, and the top 10 most frequent entity labels.

\section{Retrieval-Based Interleaved Visual CoT}
\label{sec:method}

In this section, we introduce \textit{RIV-CoT} (Retrieval-based Interleaved Visual Chain-of-Thought), a framework designed to enhance the visual reasoning abilities of VLMs by explicitly grounding their chain-of-thought process in visual evidence. RIV-CoT relies on two distinct modules: (1) the data augmentation of human-annotated explanations by interleaving them with relevant entities, and (2) a multi-step retrieval-based generation during inference to dynamically insert image crops containing relevant entities detected by the VLM into its generated chain-of-thought.

\subsection{Interleaved Explanation Augmentation}
\label{subsec:interleave} 

\begin{figure}
    \centering
    \includegraphics[width=\linewidth]{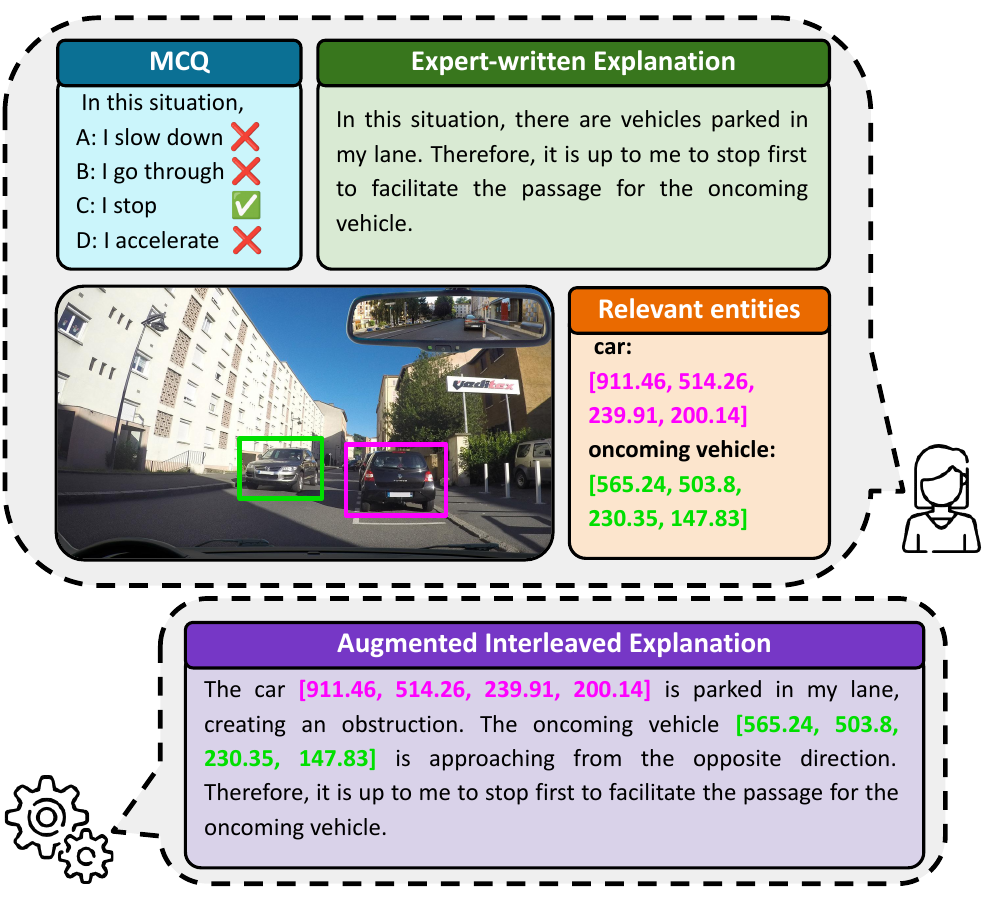}
    \caption{\textbf{Interleaved explanation augmentation.} We feed GPT-4o with the question and possible answers, the image, the list of relevant entities and coordinates, and the original expert-written explanation. The resulting interleaved explanation refers to the relevant entities early in the sentences, allowing the reasoning process to be conditioned on the content of the image crops.}
    \label{fig:interleaved_example}
\end{figure}

Explanations in \textsc{DrivingVQA} do not systematically make explicit reference to the human-annotated relevant entities. To ground explanations with visual cues, we integrate the list of relevant entities into the explanations using GPT-4o with two demonstrations, creating \emph{interleaved} explanations. 
Each set of bounding-box coordinates is embedded into the explanation, reformulating it slightly to refer to the entity at the beginning of each step of the reasoning; thereby achieving a more causal formulation aligned with the autoregressive nature of the training objective (see example in \cref{fig:interleaved_example}). This process is followed by light rule-based post-processing to refine the final explanation (see details and prompt in \cref{appx_subsec:interleaved_explanations}). 

\subsection{Multi-Step Retrieval-Based Generation}
\label{subsec:multistep-generation} 

Leveraging the interleaved augmented explanations, we propose to train a VLM using a multi-turn conversation format where an explanation is broken down into sequential turns, with each model turn ending with the bounding box coordinates of a relevant entity. To capture additional context and better understand the entity’s attributes, we expand the detected bounding box by 50\%. The expanded image crop is extracted from the input image and given as input in the next turn, allowing the model to iteratively process visual and textual information in an interleaved fashion. Training is conducted using a standard autoregressive objective.

At inference, the generation process follows an iterative approach (see \cref{fig:model_figure}). The model generates outputs until it predicts a bounding box; at that point, the process pauses and the image crop corresponding to the predicted coordinates, expanded by 50\%, is extracted. The cropped image is then encoded via the vision encoder and the patch tokens are inserted into the context after the bounding box, guiding the subsequent text generation. This multi-step retrieval and integration cycle continues until the final answer is produced, yielding intermediate steps that interleave image crops with textual reasoning, providing a more grounded CoT.

\begin{figure}[t]
  \centering
    \includegraphics[width=0.95\linewidth]{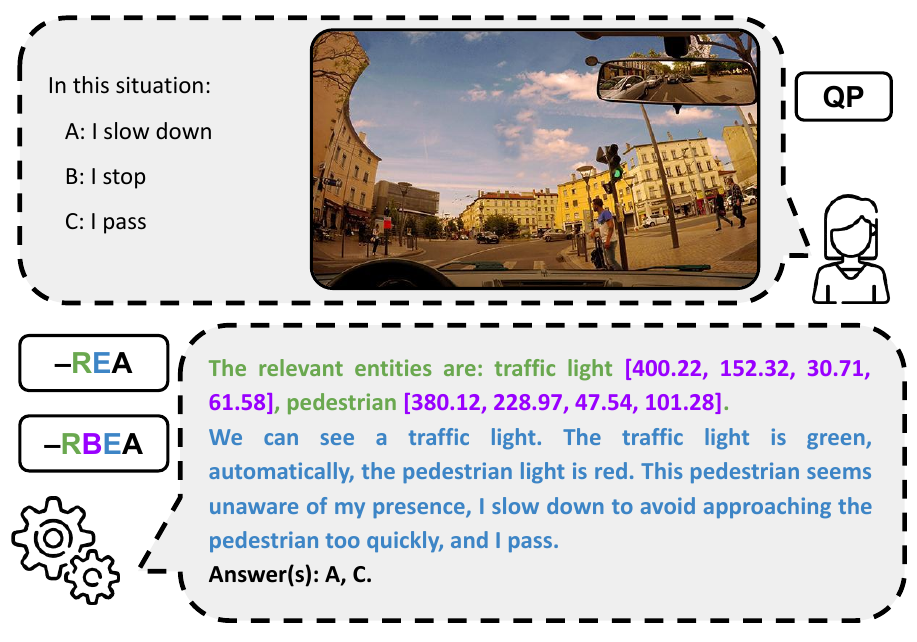}
    \caption{\textbf{Conversation formats for different fine-tuning strategies}. The model is given an image, a question and possible answer choices and is asked to predict relevant entities labels (\textbf{QP-\textcolor{ForestGreen}{R}\textcolor{RoyalBlue}{E}A}), or to predict labels and bounding box coordinates, before reasoning and answering (\textbf{QP-\textcolor{ForestGreen}{R}\textcolor{Fuchsia}{B}\textcolor{RoyalBlue}{E}A}).}
    \label{fig:conversation_format}
\end{figure}

\section{Experiments}
\label{sec:experiments}

In this section, we first evaluate the zero-shot performance of state-of-the-art VLMs on \textsc{DrivingVQA} (\cref{subsec:zero-shot}). Then we evaluate RIV-CoT, exploring the benefits of incorporating entity-related information—such as entity names, spatial coordinates, and visual content—through supervised fine-tuning to enhance the model's reasoning abilities (\cref{subsec:learning_to_generate}). We complement these results with a fine-grained analysis of the quality of the generated entity coordinates and reasoning to gain insights into possible model limitations (\cref{subsec:analysis}). Finally, we extend our study to scenarios where high-quality entity annotations are unavailable, demonstrating how automatically generated pseudo-annotations can serve as an alternative for improving model performance at scale (\cref{subsec:scaling}).

\subsection{Experimental Setup}
\label{subsec:setup}

With the exception of the zero-shot evaluation, all experiments involve fine-tuning the 7B version of LLaVA-OneVision (LLaVA-OV) \cite{li2024llava} on the \textsc{DrivingVQA} train split. LLaVA-OV uses SigLIP \cite{zhai2023sigmoid} as image encoder, a two-layer MLP projector, and Qwen2~\cite{wang2024qwen2} as LLM backbone. All components are trained end-to-end with an autoregressive loss. Each training run is repeated five times with different random seeds to account for stochasticity in the fine-tuning process and all results are reported with standard deviation. Training is performed over 10 epochs, following hyperparameters used by the authors of LLaVA-OV~\cite{li2024llava}. \\

\noindent
\textbf{Evaluation metrics.} The performance on the \textsc{DrivingVQA} test split is measured with the \textit{exam score}, analogous to the real driving theory score used to evaluate candidates: all correct answers must be selected to score a question correctly. Note that this score is equivalent to the exact match ratio. Since the task is analogous to multi-label classification, we also report the F1-Score. \\ 

\noindent
\textbf{Conversation format.} To ease the understanding of what information is provided in the prompt and what the model generates, we use the following notation based on the components of \textsc{DrivingVQA} introduced in \cref{sec:dataset}. All conversations begin with a prompt including at least the image, the question (\textbf{Q}) and the list of possible answers (\textbf{P}), and end with the predicted answer (\textbf{A}), potentially preceded by an explanation \textbf{(E)}. Hence, fine-tuning to directly answer the question is denoted as \textbf{DirectAnswer} (QP$-$A) while reasoning with the explanation before answering corresponds to \textbf{CoT} (QP$-$EA), where the dash (\textbf{$-$}) separates the input from the output of the VLM. Additional elements may include the list of the relevant entity labels (\textbf{R}), their bounding box coordinates (\textbf{B}), or the associated visual crops (\textbf{V}). The usage of interleaved explanation is denoted by \textbf{I}. Illustrative examples of conversation formats with relevant entities can be found in \cref{fig:conversation_format}, while \cref{fig:interleaved_example}'s augmented interleaved explanation showcases the IBEA format. Given this conversation format, \textbf{RIV-CoT} could also be denoted as QP$-$(IB$-$V)$^N-$EA, as it progressively incorporates multiple visual patches into the interleaved explanation across multiple steps.

\subsection{Zero-Shot Evaluation}
\label{subsec:zero-shot}

\begin{figure}[t]
    \centering
    \includegraphics[width=\linewidth]{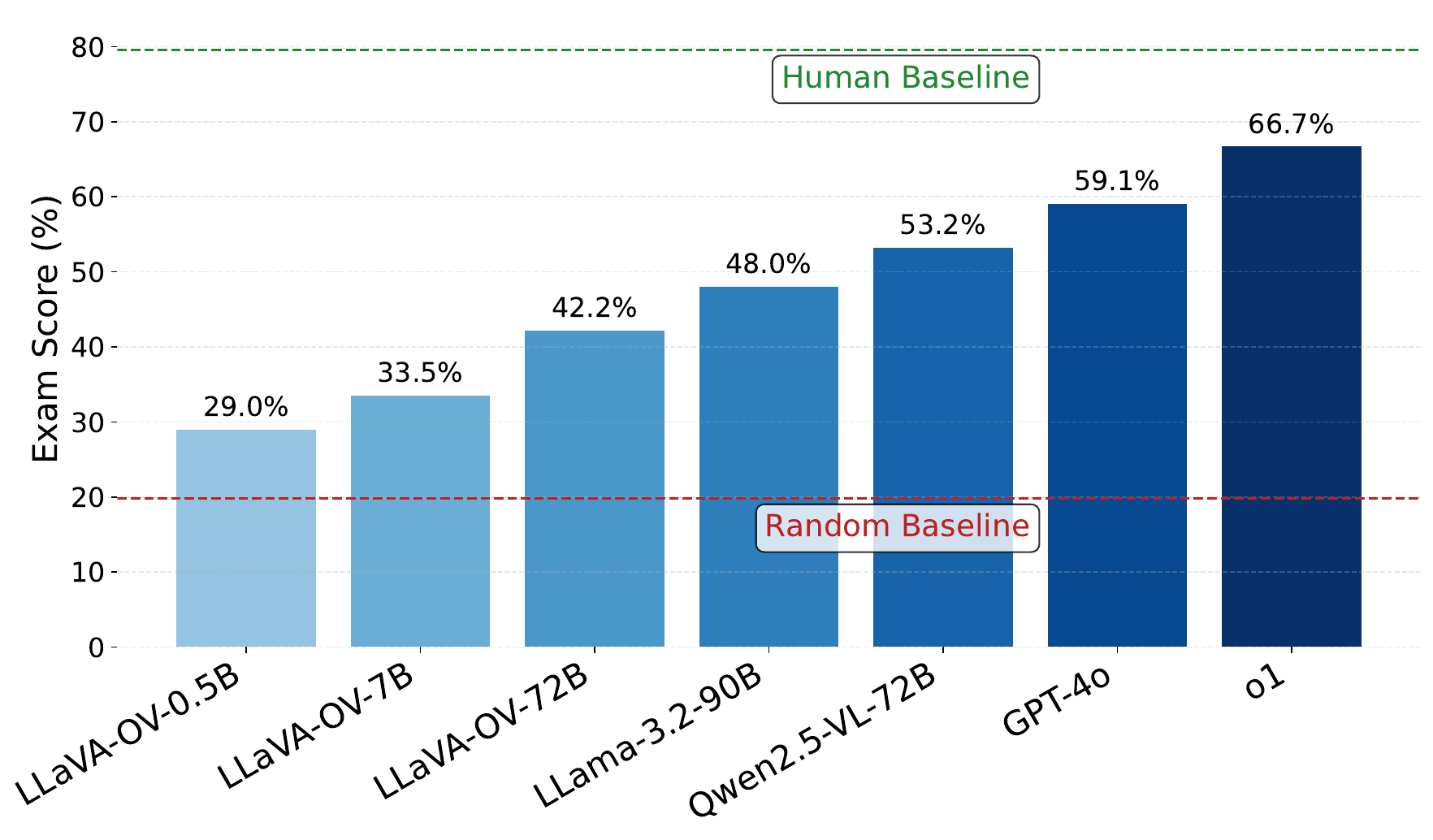}     
    \caption{\textbf{Comparison of zero-shot model performance on the \textsc{DrivingVQA} test set}. The dashed lines indicate the \textcolor{darkred}{random baseline} (19.8\%) and the \textcolor{darkgreen}{human baseline} (79.6\%).}
    \label{fig:zero-shot-results}
\end{figure}

In this section, we evaluate the zero-shot performance of popular VLMs on the \textsc{DrivingVQA} test set. For comparison, we include a random baseline where responses are selected randomly from all possible answer combinations for each question. Results are aggregated over 1,000 runs, reporting the mean exam and F1 scores. To estimate human performance, we recruit six participants with varying driving experience. They are asked to answer batches of 40 randomly selected samples in under 20 minutes, simulating the operational driving theory exam conditions. Their average exam score of 79.6\% falls below the official passing score of 87.5\% (35/40). This drop reflects the increased difficulty of \textsc{DrivingVQA}, as it focuses on visually challenging questions after the data filtering process.

\cref{fig:zero-shot-results} presents the zero-shot performance of various open-sourced models, including the LLaVA-OV variants (0.5B, 7B, 72B) used in the subsequent experiments, and proprietary models using DirectAnswer prompting. Larger models tend to perform better, with o1~\cite{openai2024openaio1card} achieving the highest exam score (66.7\%). Nevertheless, it remains far from the human baseline (79.6\%), which illustrates the benchmark’s difficulty due to its domain-specific images and knowledge and the complexity of the visual scenes. We also assess the impact of visual inputs in \textsc{DrivingVQA} by evaluating GPT-4o~\cite{openai2024gpt4o} without image information. This results in a substantial performance drop to 33.1\% (-26 pts) compared to its image-enabled counterpart. It confirms that images are necessary to accurately answer \textsc{DrivingVQA}'s questions, and knowledge-based shortcuts are not sufficient.

\begin{table}[t]
    \centering
    \setlength{\tabcolsep}{3pt} 
    \renewcommand{\arraystretch}{1.2} 
    \resizebox{\linewidth}{!}{%
    \begin{tabular}{llcccccrr}
        \toprule
        \multirow{2}{*}{\makecell[tl]{\textbf{Training} \\ \textbf{Strategy}}}   & \multicolumn{4}{c}{\textbf{Entity Format}} & \multicolumn{2}{c}{\textbf{Scores} (\%)} \\
        \cmidrule(lr){2-5} \cmidrule(lr){6-7}
         & Name & BBox & Vis. & Interlvd. & Exam & F1-Score \\
        \midrule
         DirectAnswer & & & & & 53.0 {\scriptsize \textcolor{gray}{(\(\pm\)0.9)}} & 63.3 {\scriptsize \textcolor{gray}{(\(\pm\)0.6)}} \\
         CoT &  & & & & 56.2 {\scriptsize \textcolor{gray}{(\(\pm\)1.0)}} & 65.8 {\scriptsize \textcolor{gray}{(\(\pm\)0.9)}} \\
         \midrule
         QP$-$REA & $\checkmark$ & & & & 57.0 {\scriptsize \textcolor{gray}{(\(\pm\)1.3)}} & 67.1 {\scriptsize \textcolor{gray}{(\(\pm\)1.4)}} \\
         QP$-$RBEA & $\checkmark$ & $\checkmark$ & & & 57.7 {\scriptsize \textcolor{gray}{(\(\pm\)0.7)}} & 67.3 {\scriptsize \textcolor{gray}{(\(\pm\)0.7)}} \\
         QP$-$RB$-$RV$-$EA & $\checkmark$ & $\checkmark$ & $\checkmark$ & & 58.4 {\scriptsize \textcolor{gray}{(\(\pm\)1.1)}} & 67.8 {\scriptsize \textcolor{gray}{(\(\pm\)1.1)}} \\
         \midrule
         QP$-$IEA & $\checkmark$ & $\checkmark$ & & $\checkmark$ & 56.4 {\scriptsize \textcolor{gray}{(\(\pm\)0.4)}} & 66.3 {\scriptsize \textcolor{gray}{(\(\pm\)0.6)}} \\
         QP$-$IBEA & $\checkmark$ & $\checkmark$ & & $\checkmark$ & 57.9 {\scriptsize \textcolor{gray}{(\(\pm\)0.5)}} & 66.8 {\scriptsize \textcolor{gray}{(\(\pm\)0.2)}} \\
         \rowcolor{gray!25} RIV-CoT & $\checkmark$ & $\checkmark$ & $\checkmark$ & $\checkmark$ & \bfseries 59.3 {\scriptsize \textcolor{gray}{(\(\pm\)1.0)}} & \bfseries 68.8 {\scriptsize \textcolor{gray}{(\(\pm\)0.9)}} \\    
        \bottomrule
    \end{tabular}
    }
    \caption{\textbf{Comparative results of fine-tuning LLaVA-OV-7B using different training strategies}. \textit{Vis.} stands for ``visual crops''. 
    }
\label{tab:model_performance_fine_tuning}
\end{table}

\subsection{Fine-tuning with Relevant Entities}
\label{subsec:learning_to_generate}

We explore how to enhance VLMs' visual chain-of-thought by leveraging relevant entities through their label, bounding box and visual crops, all of these interleaved in the explanation or not (\cref{tab:model_performance_fine_tuning}); demonstrating the importance of each component of RIV-CoT.

First, we experiment with the \textbf{DirectAnswer} and chain-of-thought prompting (\textbf{CoT}) baselines. As expected, fine-tuning LLaVA-OV with \textbf{CoT} allows the model to generate better answers than predicting answers directly. 
While learning to generate the list of relevant entities (\textbf{QP$-$REA}) before generating an explanation and answer only leads to a minor improvement, adding bounding box coordinates (\textbf{QP$-$RBEA}) improves the exam score to 57.7\% (+1.5 pts over \textbf{CoT}). 
Then, to leverage the visual content, we use a two-step reasoning format (\textbf{QP$-$RB$-$RV$-$EA}), where the model first predicts relevant entities' labels and coordinates given the question and answer choices (QP$-$RB). Then, it is fed with the predicted entities' visual patches,  and uses it to reason and answer ($-$RV$-$EA).\footnote{Note that the training strategy of \textbf{QP$-$RB$-$RV$-$EA} shares similarities with Visual CoT \cite{shao2024visual} but uses a different visual patch cropping strategy and makes use of an unlimited number of relevant entities per sample.}
This strategy further improves the performance, reaching 58.4\% and showing the importance of visual patches for reasoning.

Finally, using explanations interleaved with visual patches, \textbf{RIV-CoT} achieves the best exam score (59.3\%). This result shows that providing rich contextual visual information interleaved within explanations results in the most efficient way to enhance VLMs' reasoning abilities. Fine-tuning with interleaved explanation without coordinates nor visual patches (\textbf{QP$-$IEA}) leads to a score comparable to fine-tuning with the original explanations (\textbf{CoT}), validating that the performance gain seen in \textbf{QP$-$IBEA} and \textbf{RIV-CoT} comes from the added information from the bounding boxes and visual patches.

\subsection{Analysis}
\label{subsec:analysis}

We analyze the impact of the correctness of relevant entities' predicted coordinates on model performance, and the link between model performance, reasoning correctness and fine-tuning strategy.

\subsubsection{Impact of entity detection correctness}
The detection performance of models fine-tuned to predict bounding boxes of relevant entities is presented in \cref{tab:detection_metrics}. We measure the top-1 accuracy based on Hungarian matching at an IoU threshold of 0.50. Results are reported for all samples, as well as for samples with correct and incorrect final predictions.

\begin{table}[ht]
    \centering
    \setlength{\tabcolsep}{8pt} 
    \renewcommand{\arraystretch}{1.2} 
    \resizebox{\columnwidth}{!}{%
    \begin{tabular}{lccc}
        \toprule
        \multirow{2}{*}{\makecell[tl]{\textbf{Training} \\ \textbf{Strategy}}} & \multicolumn{3}{c}{\textbf{Top-1 Acc. @ IoU0.50 (\%)}} \\
        \cmidrule(lr){2-4}
        & \textbf{All} & \textbf{Correct} & \textbf{Incorrect} \\ 
        \midrule
        QP$-$RBEA      & 68.7 {\scriptsize \textcolor{gray}{(\(\pm\)1.4)}} & 69.3 {\scriptsize \textcolor{gray}{(\(\pm\)1.1)}} & \bfseries 67.6 {\scriptsize \textcolor{gray}{(\(\pm\)1.4)}} \\
        QP$-$IBEA      & 68.5 {\scriptsize \textcolor{gray}{(\(\pm\)1.8)}} & 68.9 {\scriptsize \textcolor{gray}{(\(\pm\)1.7)}} & 67.3 {\scriptsize \textcolor{gray}{(\(\pm\)1.6)}} \\
        QP$-$RB$-$RV$-$EA  & \bfseries 69.8 {\scriptsize \textcolor{gray}{(\(\pm\)1.5)}} & 72.3 {\scriptsize \textcolor{gray}{(\(\pm\)1.4)}} & 66.1 {\scriptsize \textcolor{gray}{(\(\pm\)2.7)}} \\
        RIV-CoT   & 69.6 {\scriptsize \textcolor{gray}{(\(\pm\)1.7)}} & \bfseries 72.4 {\scriptsize \textcolor{gray}{(\(\pm\)1.2)}} &  66.3 {\scriptsize \textcolor{gray}{(\(\pm\)1.8)}} \\
        \bottomrule
    \end{tabular}
    }
    \caption{\textbf{Detection performance of bounding box-predicting models.} Top-1 accuracy at an IoU threshold of 50\% is reported for all samples, correctly predicted, and incorrectly predicted samples.}
    \label{tab:detection_metrics}
\end{table}

For correctly answered samples, the detection accuracy is slightly higher than the overall average; while samples with incorrect predictions exhibit lower accuracy. Notably, \textbf{RIV-CoT} achieves the highest accuracy on correctly predicted samples (72.4\%) but its detection performance decreases to 66.3\% on incorrect answers, indicating that detection failures may contribute to prediction errors. 

\begin{table}[t]
    \centering
    \setlength{\tabcolsep}{7pt} 
    \renewcommand{\arraystretch}{1.3} 
    \resizebox{\columnwidth}{!}{%
    \begin{tabular}{lcc}
        \toprule
        \textbf{Visual Patch Type} & \textbf{Exam (\%)} & \textbf{F1-Score (\%)} \\
        \midrule
        Predicted Entities (QP$-$RB$-$RV$-$EA)  & 58.7 {\scriptsize \textcolor{gray}{(\(\pm\)0.8)}} & 67.9 {\scriptsize \textcolor{gray}{(\(\pm\)1.1)}} \\
        Image Split (QPR$-$EA\(^{\dagger}\))  & 60.7 {\scriptsize \textcolor{gray}{(\(\pm\)0.9)}} & 69.8 {\scriptsize \textcolor{gray}{(\(\pm\)0.6)}} \\
        Oracle Entities (QPRV$-$EA)         & \bfseries 62.0 {\scriptsize \textcolor{gray}{(\(\pm\)0.4)}} & \bfseries 71.0 {\scriptsize \textcolor{gray}{(\(\pm\)0.5)}} \\
        \bottomrule
    \end{tabular}
    }
    \caption{\textbf{Performance analysis by type of visual patches.} Results are shown for models using predicted entities, oracle entities, and a multi-scale image splitting strategy. $\dagger$ indicates a model trained with AnyRes-4~\cite{liu2024llavanext} visual patches.}
    \label{tab:quality_patches}
\end{table}

To further assess the effectiveness of visual patches, we compare models using predicted entities (\textbf{QP$-$RB$-$RV$-$EA}) against oracle entities (\textbf{QPRV$-$EA}) in \cref{tab:quality_patches}. The oracle entities improve performance significantly (+3.3 pts in exam score, +3.1 pts in F1-score), highlighting a limitation in grounding capabilities of the VLM chosen in our experiment.

To determine whether this improvement stems from relevant visual information rather than merely more visual tokens, we compare \textbf{QPRV$-$EA} with a multi-scale image patching strategy (\textbf{QPR$-$EA$^{\dagger}$}), called \textit{AnyRes}~\cite{liu2024llavanext}. It consists in splitting the image into $N$ sub-patches and concatenating all their representations with the original image, and is often used to encode high-resolution images. As the maximum number of entities per sample in \textsc{DrivingVQA} is five, we use $N=4$ for a fair comparison. Results in \cref{tab:quality_patches} show that \textbf{QPR$-$EA$^{\dagger}$} only achieve 60.7\% exam score, -1.3 pts compared to \textbf{QPRV$-$EA}, which confirms that carefully selected visual patches provide a stronger advantage than generic image splitting.

\begin{figure}[t]
    \centering
    \includegraphics[width=\linewidth]{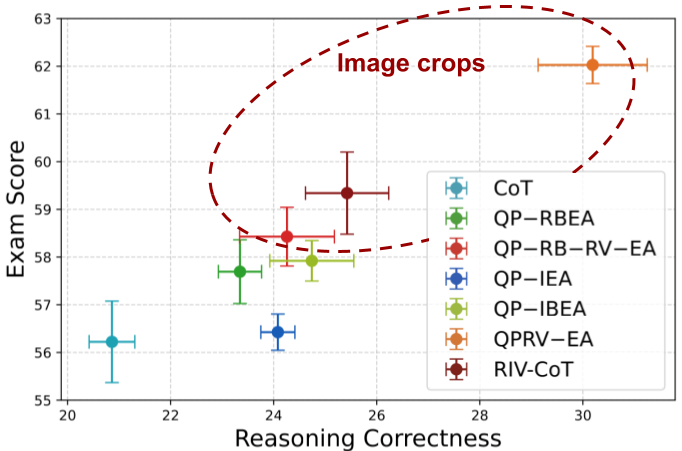}
    \caption{\textbf{Reasoning correctness vs. Exam score for each fine-tuning strategy.} In shades of green are methods using bounding boxes; in shades of red, using image crops. 
    }
    \label{fig:reasoning-correctness}
\end{figure}

\subsubsection{Reasoning Correctness}

Our evaluation metrics, the exam score and the F1-score, measure a model’s ability to select the correct answer. However, our approach is designed to enhance the model's VQA performance by making its reasoning more grounded in the input image. While a high exam score indicates an improved answer selection, it does not necessarily reflect the correctness of the reasoning. A model may rely on heuristics or memorization, a limitation that the exam score alone does not fully capture. To address this, we assess the correctness of the model's reasoning by comparing it against \textsc{DrivingVQA}'s ground truth explanations. We use GPT-4o-mini as an evaluator, leveraging the LLM-as-a-judge paradigm~\cite{zheng2023judging}, which is increasingly used to scale the evaluation of open-ended generations of LLMs and VLMs, and is shown to align well with human judgment in pairwise comparisons \cite{liu2024aligning, gu2024survey}.

The prompt used for GPT-4o-mini is detailed in \cref{appx_subsec:reasoning_eval}. The model is provided with the question and answer choices, the ground truth explanation, and the reasoning generated by the VLM. It is then tasked with identifying the key arguments in both reasonings, checking for missing or contradictory elements, and determining whether they align. On a balanced set of 50 human-evaluated reasoning pairs sampled from different fine-tuning strategies, GPT-4o-mini achieves the same judging performance as GPT-4o, with an F1-score of 0.82. We define the \textit{reasoning correctness} as the proportion of samples in the test set where the judge determines that the model’s reasoning matches the ground truth reasoning. 

We compare reasoning correctness and exam scores across our systems in \cref{fig:reasoning-correctness}. While the two metrics are positively correlated, the reasoning correctness consistently remains at least half of the exam score. The \textbf{CoT} baseline reaches a reasoning correctness of 20.8\%; then come models incorporating bounding boxes (\textbf{QP$-$RBEA}, \textbf{QP$-$IBVEA}) and image crops (\textbf{QP$-$RB$-$RV$-$EA}, \textbf{RIV-CoT}). As a comparison, \textbf{QPRV$-$EA}, which receives image crops of oracle entities as input, achieves 30.2\% reasoning correctness. Among retrieval-based approaches, \textbf{RIV-CoT} outperforms its non-interleaved counterpart \textbf{QP$-$RB$-$RV$-$EA} by 1.2 points and outperforms the \textbf{CoT} baseline by 4.6 points, demonstrating the benefits of visual patches and interleaved formatting for reasoning accuracy.

\subsection{Scaling with Automatically-Extracted Entities}
\label{subsec:scaling}

\textsc{DrivingVQA} includes high-quality annotated relevant entities along with their coordinates. Such annotations are rarely available, and transferring our method to new datasets requires external tools to obtain annotations. In this section, we extend our approach to A-OKVQA~\cite{aokvqa2022}, a crowd-sourced dataset composed of 25K visual questions requiring commonsense and world knowledge to be answered. Similar to \textsc{DrivingVQA}, each question in A-OKVQA is accompanied by multiple-choice options and rationales explaining the reasoning behind the correct answer. To obtain a list of relevant entities and their coordinates for each question, we employ an automated entity extraction pipeline based on the method introduced in \cref{subsec:relevant_entities}. Specifically, we prompt GPT-4o-mini to generate potential relevant entity labels for each question. Then, we use the open-set object detector GroundingDINO~\cite{liu2023grounding} to localize these entities within the image. Only the top 5 entities with the highest confidence scores are retained. 

\begin{table}[t]
    \centering
    \setlength{\tabcolsep}{8pt}
    \renewcommand{\arraystretch}{1.2}
    \resizebox{0.9\linewidth}{!}{%
    \begin{tabular}{llc}
        \toprule
        \textbf{Dataset} & \textbf{Training Strategy} & \textbf{MC Acc. (\%)} \\
        \midrule
        \multirow{4}{*}{A-OKVQA}  
         & DirectAnswer     & 78.2 {\scriptsize \textcolor{gray}{(\(\pm\)0.3)}}  \\
         & CoT              & 80.6 {\scriptsize \textcolor{gray}{(\(\pm\)0.4)}} \\
         & QP$-$RB$-$RV$-$CoT      & 82.3 {\scriptsize \textcolor{gray}{(\(\pm\)0.3)}} \\
         & RIV-CoT          & \bfseries 84.2 {\scriptsize \textcolor{gray}{(\(\pm\)0.2)}}\\    
        \bottomrule
    \end{tabular}
    }
    \caption{\textbf{Multiple-choice accuracy on the val subset of A-OKVQA~\cite{aokvqa2022}}, for different training strategies.}
\label{tab:aokvqa-results}
\end{table}

Using these pseudo-labeled entities, we fine-tune LLaVA-OV-7B with our \textbf{RIV-CoT} method on the train set of A-OKVQA (17K samples), and evaluate its performance on the available validation subset (1992 samples). As shown in \cref{tab:aokvqa-results}, \textbf{RIV-CoT} achieves a multiple-choice accuracy of 84.2\%, outperforming both the \textbf{DirectAnswer} baseline (78.2\%) and vanilla \textbf{CoT} prompting (80.6\%). Notably, the two-step conversation strategy \textbf{QP$-$RB$-$RV$-$EA}, which integrates visual patches without interleaving them through retrieval-based generation, achieves 82.3\% accuracy. These results highlight that our retrieval-based interleaved approach can largely improve the model’s ability to accurately answer questions that require complex visual reasoning, even when relying on automatically extracted entities to compensate for the lack of gold-standard annotations. 

\section{Conclusion}
\label{sec:conclusion}

In this work, we introduce \textsc{DrivingVQA}, a novel visual reasoning dataset derived from French driving theory exams, along with RIV-CoT, a retrieval-based interleaved visual chain-of-thought framework designed to enhance the visual reasoning abilities of VLMs. Our experiments on \textsc{DrivingVQA} demonstrate that RIV-CoT significantly improves both answer accuracy and reasoning correctness compared to direct answer generation and vanilla chain-of-thought prompting. Furthermore, we show that our approach scales effectively to datasets lacking human annotations by leveraging automatically generated pseudo-labels, ensuring broader applicability in real-world scenarios. 

Despite these promising advances, our analysis highlights challenges that VLMs face in zero-shot settings. A promising direction for future research is to integrate retrieval-based interleaved visual chain-of-thought reasoning into VLM's pretraining, by scaling the automatically extracted entities variant on extensive grounding datasets and established multimodal reasoning benchmarks.

\section*{Acknowledgments}
We thank Max Luca Pio Conti, Pierre Ancey, Francesco Pettenon and Matthias Wyss for their contributions to preliminary work. We thank Auguste Poiroux, Gaston Lenczner, Florent Forest, Jacques Everwyn, Vincent Montariol, Alice Legrand, Marc Lafon, Yannis Karmim, and Alexandre Merkli for the human evaluation of the \textsc{DrivingVQA} test set. We also thank the VITA
lab members for their valuable feedback, which helped to
enhance the quality of this manuscript.
SM gratefully acknowledges the support of the Swiss National Science Foundation (No. 224881). AB gratefully acknowledges the support of the Swiss National Science Foundation (No. 215390), Innosuisse (PFFS-21-29), the EPFL Center for Imaging, Sony Group Corporation, and the Allen Institute for AI.

{
    \small
    \bibliographystyle{ieeenat_fullname}
    \bibliography{main}

\begin{thebibliography}{59}
\providecommand{\natexlab}[1]{#1}
\providecommand{\url}[1]{\texttt{#1}}
\expandafter\ifx\csname urlstyle\endcsname\relax
  \providecommand{\doi}[1]{doi: #1}\else
  \providecommand{\doi}{doi: \begingroup \urlstyle{rm}\Url}\fi

\bibitem[Alayrac et~al.(2022)Alayrac, Donahue, Luc, Miech, Barr, Hasson, Lenc, Mensch, Millican, Reynolds, et~al.]{alayrac2022flamingo}
Jean-Baptiste Alayrac, Jeff Donahue, Pauline Luc, Antoine Miech, Iain Barr, Yana Hasson, Karel Lenc, Arthur Mensch, Katherine Millican, Malcolm Reynolds, et~al.
\newblock Flamingo: a visual language model for few-shot learning.
\newblock \emph{Advances in neural information processing systems}, 35:\penalty0 23716--23736, 2022.

\bibitem[Antol et~al.(2015)Antol, Agrawal, Lu, Mitchell, Batra, Zitnick, and Parikh]{VQA2015}
Stanislaw Antol, Aishwarya Agrawal, Jiasen Lu, Margaret Mitchell, Dhruv Batra, C.~Lawrence Zitnick, and Devi Parikh.
\newblock {VQA}: {V}isual {Q}uestion {A}nswering.
\newblock In \emph{International Conference on Computer Vision (ICCV)}, 2015.

\bibitem[Chandhok et~al.(2024)Chandhok, Fan, and Sigal]{chandhok2024response}
Shivam Chandhok, Wan-Cyuan Fan, and Leonid Sigal.
\newblock Response wide shut: Surprising observations in basic vision language model capabilities.
\newblock \emph{arXiv preprint arXiv:2408.06721}, 2024.

\bibitem[Chen et~al.(2023)Chen, Zhang, Zeng, Zhang, Zhu, and Zhao]{chen2023shikra}
Keqin Chen, Zhao Zhang, Weili Zeng, Richong Zhang, Feng Zhu, and Rui Zhao.
\newblock Shikra: Unleashing multimodal llm's referential dialogue magic.
\newblock \emph{arXiv preprint arXiv:2306.15195}, 2023.

\bibitem[Chia et~al.(2024)Chia, Han, Ghosal, Bing, and Poria]{chia2024puzzlevqa}
Yew~Ken Chia, Vernon Toh~Yan Han, Deepanway Ghosal, Lidong Bing, and Soujanya Poria.
\newblock Puzzlevqa: Diagnosing multimodal reasoning challenges of language models with abstract visual patterns.
\newblock \emph{arXiv preprint arXiv:2403.13315}, 2024.

\bibitem[Contributors(2024)]{openai2024gpt4technicalreport}
OpenAI Contributors.
\newblock Gpt-4 technical report, 2024.

\bibitem[Goyal et~al.(2017)Goyal, Khot, Summers{-}Stay, Batra, and Parikh]{GoyalKSBP16}
Yash Goyal, Tejas Khot, Douglas Summers{-}Stay, Dhruv Batra, and Devi Parikh.
\newblock Making the {V} in {VQA} matter: Elevating the role of image understanding in visual question answering.
\newblock In \emph{Proceedings of the IEEE/CVF Conference on Computer Vision and Pattern Recognition (CVPR)}, 2017.

\bibitem[Gu et~al.(2024)Gu, Jiang, Shi, Tan, Zhai, Xu, Li, Shen, Ma, Liu, et~al.]{gu2024survey}
Jiawei Gu, Xuhui Jiang, Zhichao Shi, Hexiang Tan, Xuehao Zhai, Chengjin Xu, Wei Li, Yinghan Shen, Shengjie Ma, Honghao Liu, et~al.
\newblock A survey on llm-as-a-judge.
\newblock \emph{arXiv preprint arXiv:2411.15594}, 2024.

\bibitem[Guan et~al.(2024)Guan, Liu, Wu, Xian, Li, Liu, Wang, Chen, Huang, Yacoob, Manocha, and Zhou]{Guan_2024_CVPR}
Tianrui Guan, Fuxiao Liu, Xiyang Wu, Ruiqi Xian, Zongxia Li, Xiaoyu Liu, Xijun Wang, Lichang Chen, Furong Huang, Yaser Yacoob, Dinesh Manocha, and Tianyi Zhou.
\newblock Hallusionbench: An advanced diagnostic suite for entangled language hallucination and visual illusion in large vision-language models.
\newblock In \emph{Proceedings of the IEEE/CVF Conference on Computer Vision and Pattern Recognition (CVPR)}, pages 14375--14385, 2024.

\bibitem[Gupta and Kembhavi(2023)]{gupta2023visual}
Tanmay Gupta and Aniruddha Kembhavi.
\newblock Visual programming: Compositional visual reasoning without training.
\newblock In \emph{Proceedings of the IEEE/CVF Conference on Computer Vision and Pattern Recognition}, pages 14953--14962, 2023.

\bibitem[Hudson and Manning(2019)]{hudson2019gqa}
Drew~A Hudson and Christopher~D Manning.
\newblock Gqa: A new dataset for real-world visual reasoning and compositional question answering.
\newblock In \emph{Proceedings of the IEEE/CVF Conference on Computer Vision and Pattern Recognition (CVPR)}, 2019.

\bibitem[Hwang et~al.(2024)Hwang, Xu, Lin, Hung, Ji, Choi, Huang, He, Covington, Sapp, Zhou, Guo, Anguelov, and Tan]{hwang2024emma}
Jyh-Jing Hwang, Runsheng Xu, Hubert Lin, Wei-Chih Hung, Jingwei Ji, Kristy Choi, Di Huang, Tong He, Paul Covington, Benjamin Sapp, Yin Zhou, James Guo, Dragomir Anguelov, and Mingxing Tan.
\newblock Emma: End-to-end multimodal model for autonomous driving.
\newblock \emph{arXiv preprint arXiv:2410.23262}, 2024.

\bibitem[Jiang et~al.(2023)Jiang, Sablayrolles, Mensch, Bamford, Chaplot, de~las Casas, Bressand, Lengyel, Lample, Saulnier, Lavaud, Lachaux, Stock, Scao, Lavril, Wang, Lacroix, and Sayed]{jiang2023mistral7b}
Albert~Q. Jiang, Alexandre Sablayrolles, Arthur Mensch, Chris Bamford, Devendra~Singh Chaplot, Diego de~las Casas, Florian Bressand, Gianna Lengyel, Guillaume Lample, Lucile Saulnier, Lélio~Renard Lavaud, Marie-Anne Lachaux, Pierre Stock, Teven~Le Scao, Thibaut Lavril, Thomas Wang, Timothée Lacroix, and William~El Sayed.
\newblock Mistral 7b, 2023.

\bibitem[Jiang et~al.(2024)Jiang, Xu, Dong, Chen, Ye, Yan, Ye, Zhang, Huang, and Zhang]{jiang2024hallucination}
Chaoya Jiang, Haiyang Xu, Mengfan Dong, Jiaxing Chen, Wei Ye, Ming Yan, Qinghao Ye, Ji Zhang, Fei Huang, and Shikun Zhang.
\newblock Hallucination augmented contrastive learning for multimodal large language model.
\newblock In \emph{Proceedings of the IEEE/CVF Conference on Computer Vision and Pattern Recognition}, pages 27036--27046, 2024.

\bibitem[Kim et~al.(2018)Kim, Rohrbach, Darrell, Canny, and Akata]{kim2018textual}
Jinkyu Kim, Anna Rohrbach, Trevor Darrell, John Canny, and Zeynep Akata.
\newblock Textual explanations for self-driving vehicles.
\newblock In \emph{European Conference on Computer Vision (ECCV)}, 2018.

\bibitem[Lei et~al.(2024)Lei, Yang, Chen, Li, and Liu]{lei2024scaffolding}
Xuanyu Lei, Zonghan Yang, Xinrui Chen, Peng Li, and Yang Liu.
\newblock Scaffolding coordinates to promote vision-language coordination in large multi-modal models.
\newblock \emph{arXiv preprint arXiv:2402.12058}, 2024.

\bibitem[Li et~al.(2024{\natexlab{a}})Li, Zhang, Guo, Zhang, Li, Zhang, Zhang, Li, Liu, and Li]{li2024llava}
Bo Li, Yuanhan Zhang, Dong Guo, Renrui Zhang, Feng Li, Hao Zhang, Kaichen Zhang, Yanwei Li, Ziwei Liu, and Chunyuan Li.
\newblock Llava-onevision: Easy visual task transfer.
\newblock \emph{arXiv preprint arXiv:2408.03326}, 2024{\natexlab{a}}.

\bibitem[Li et~al.(2022)Li, Niu, and Zhang]{li2022from}
Jiangtong Li, Li Niu, and Liqing Zhang.
\newblock From representation to reasoning: Towards both evidence and commonsense reasoning for video question-answering.
\newblock In \emph{Proceedings of the IEEE/CVF Conference on Computer Vision and Pattern Recognition (CVPR)}, 2022.

\bibitem[Li et~al.(2025)Li, Sun, Cheng, Zhu, and Chen]{li2025chain}
Xue Li, Yiyou Sun, Wei Cheng, Yinglun Zhu, and Haifeng Chen.
\newblock Chain-of-region: Visual language models need details for diagram analysis.
\newblock In \emph{The Thirteenth International Conference on Learning Representations}, 2025.

\bibitem[Li et~al.(2024{\natexlab{b}})Li, Liu, Zhang, Wang, Xue, and Cai]{li2024enhancing}
Zhiyuan Li, Dongnan Liu, Chaoyi Zhang, Heng Wang, Tengfei Xue, and Weidong Cai.
\newblock Enhancing advanced visual reasoning ability of large language models.
\newblock In \emph{Proceedings of the 2024 Conference on Empirical Methods in Natural Language Processing}, pages 1915--1929, 2024{\natexlab{b}}.

\bibitem[Lin et~al.(2024)Lin, Yin, Ping, Molchanov, Shoeybi, and Han]{lin2024vila}
Ji Lin, Hongxu Yin, Wei Ping, Pavlo Molchanov, Mohammad Shoeybi, and Song Han.
\newblock Vila: On pre-training for visual language models.
\newblock In \emph{Proceedings of the IEEE/CVF Conference on Computer Vision and Pattern Recognition}, pages 26689--26699, 2024.

\bibitem[Liu et~al.(2024{\natexlab{a}})Liu, Li, Li, Li, Zhang, Shen, and Lee]{liu2024llavanext}
Haotian Liu, Chunyuan Li, Yuheng Li, Bo Li, Yuanhan Zhang, Sheng Shen, and Yong~Jae Lee.
\newblock Llava-next: Improved reasoning, ocr, and world knowledge, 2024{\natexlab{a}}.

\bibitem[Liu et~al.(2024{\natexlab{b}})Liu, Li, Wu, and Lee]{liu2024visual}
Haotian Liu, Chunyuan Li, Qingyang Wu, and Yong~Jae Lee.
\newblock Visual instruction tuning.
\newblock \emph{Advances in Neural Information Processing Systems (NeurIPS)}, 2024{\natexlab{b}}.

\bibitem[Liu et~al.(2024{\natexlab{c}})Liu, Xue, Chen, Chen, Zhao, Wang, Hou, Li, and Peng]{liu2024survey}
Hanchao Liu, Wenyuan Xue, Yifei Chen, Dapeng Chen, Xiutian Zhao, Ke Wang, Liping Hou, Rongjun Li, and Wei Peng.
\newblock A survey on hallucination in large vision-language models.
\newblock \emph{arXiv preprint arXiv:2402.00253}, 2024{\natexlab{c}}.

\bibitem[Liu et~al.(2023)Liu, Zeng, Ren, Li, Zhang, Yang, Jiang, Li, Yang, Su, et~al.]{liu2023grounding}
Shilong Liu, Zhaoyang Zeng, Tianhe Ren, Feng Li, Hao Zhang, Jie Yang, Qing Jiang, Chunyuan Li, Jianwei Yang, Hang Su, et~al.
\newblock Grounding dino: Marrying dino with grounded pre-training for open-set object detection.
\newblock \emph{arXiv preprint arXiv:2303.05499}, 2023.

\bibitem[Liu et~al.(2024{\natexlab{d}})Liu, Zhou, Guo, Shareghi, Vuli{\'c}, Korhonen, and Collier]{liu2024aligning}
Yinhong Liu, Han Zhou, Zhijiang Guo, Ehsan Shareghi, Ivan Vuli{\'c}, Anna Korhonen, and Nigel Collier.
\newblock Aligning with human judgement: The role of pairwise preference in large language model evaluators.
\newblock \emph{arXiv preprint arXiv:2403.16950}, 2024{\natexlab{d}}.

\bibitem[Liu et~al.(2025{\natexlab{a}})Liu, Chi, Wu, Zhang, Hu, Zhang, Zhang, Wu, Cao, Huang, et~al.]{liu2025spatialcot}
Yuecheng Liu, Dafeng Chi, Shiguang Wu, Zhanguang Zhang, Yaochen Hu, Lingfeng Zhang, Yingxue Zhang, Shuang Wu, Tongtong Cao, Guowei Huang, et~al.
\newblock Spatialcot: Advancing spatial reasoning through coordinate alignment and chain-of-thought for embodied task planning.
\newblock \emph{arXiv preprint arXiv:2501.10074}, 2025{\natexlab{a}}.

\bibitem[Liu et~al.(2025{\natexlab{b}})Liu, Duan, Zhang, Li, Zhang, Zhao, Yuan, Wang, He, Liu, et~al.]{liu2025mmbench}
Yuan Liu, Haodong Duan, Yuanhan Zhang, Bo Li, Songyang Zhang, Wangbo Zhao, Yike Yuan, Jiaqi Wang, Conghui He, Ziwei Liu, et~al.
\newblock Mmbench: Is your multi-modal model an all-around player?
\newblock In \emph{European Conference on Computer Vision}, pages 216--233. Springer, 2025{\natexlab{b}}.

\bibitem[Lu et~al.(2022)Lu, Mishra, Xia, Qiu, Chang, Zhu, Tafjord, Clark, and Kalyan]{lu2022learn}
Pan Lu, Swaroop Mishra, Tanglin Xia, Liang Qiu, Kai-Wei Chang, Song-Chun Zhu, Oyvind Tafjord, Peter Clark, and Ashwin Kalyan.
\newblock Learn to explain: Multimodal reasoning via thought chains for science question answering.
\newblock \emph{Advances in Neural Information Processing Systems}, 35:\penalty0 2507--2521, 2022.

\bibitem[Marcu et~al.(2024)Marcu, Chen, Hünermann, Karnsund, Hanotte, Chidananda, Nair, Badrinarayanan, Kendall, Shotton, Arani, and Sinavski]{marcu2024lingoqa}
Ana-Maria Marcu, Long Chen, Jan Hünermann, Alice Karnsund, Benoit Hanotte, Prajwal Chidananda, Saurabh Nair, Vijay Badrinarayanan, Alex Kendall, Jamie Shotton, Elahe Arani, and Oleg Sinavski.
\newblock Lingoqa: Visual question answering for autonomous driving.
\newblock In \emph{European Conference on Computer Vision (ECCV)}, 2024.

\bibitem[Marino et~al.(2019)Marino, Rastegari, Farhadi, and Mottaghi]{okvqa2019}
Kenneth Marino, Mohammad Rastegari, Ali Farhadi, and Roozbeh Mottaghi.
\newblock Ok-vqa: A visual question answering benchmark requiring external knowledge.
\newblock In \emph{Proceedings of the IEEE/CVF Conference on Computer Vision and Pattern Recognition (CVPR)}, 2019.

\bibitem[Mitra et~al.(2024)Mitra, Huang, Darrell, and Herzig]{mitra2024compositional}
Chancharik Mitra, Brandon Huang, Trevor Darrell, and Roei Herzig.
\newblock Compositional chain-of-thought prompting for large multimodal models.
\newblock In \emph{Proceedings of the IEEE/CVF Conference on Computer Vision and Pattern Recognition}, pages 14420--14431, 2024.

\bibitem[OpenAI(2024{\natexlab{a}})]{openai2024gpt4o}
OpenAI.
\newblock Gpt-4o system card, 2024{\natexlab{a}}.

\bibitem[OpenAI(2024{\natexlab{b}})]{openai2024openaio1card}
OpenAI.
\newblock Openai o1 system card, 2024{\natexlab{b}}.

\bibitem[{\"O}zdemir and Akag{\"u}nd{\"u}z(2024)]{ozdemir2024enhancing}
{\"O}vg{\"u} {\"O}zdemir and Erdem Akag{\"u}nd{\"u}z.
\newblock Enhancing visual question answering through question-driven image captions as prompts.
\newblock In \emph{Proceedings of the IEEE/CVF Conference on Computer Vision and Pattern Recognition}, pages 1562--1571, 2024.

\bibitem[Piaget(1978)]{piaget1978piaget}
Jean Piaget.
\newblock \emph{Piaget's theory of intelligence}.
\newblock Englewood Cliffs, NJ: Prentice Hall, 1978.

\bibitem[Qi et~al.(2024)Qi, Ding, Wang, Bai, Lv, Hong, Xu, Hou, Li, Dong, et~al.]{qi2024cogcom}
Ji Qi, Ming Ding, Weihan Wang, Yushi Bai, Qingsong Lv, Wenyi Hong, Bin Xu, Lei Hou, Juanzi Li, Yuxiao Dong, et~al.
\newblock Cogcom: Train large vision-language models diving into details through chain of manipulations.
\newblock \emph{arXiv preprint arXiv:2402.04236}, 2024.

\bibitem[Qian et~al.(2024)Qian, Chen, Zhuo, Jiao, and Jiang]{qian2024nuscenesqa}
Tianwen Qian, Jingjing Chen, Linhai Zhuo, Yang Jiao, and Yu-Gang Jiang.
\newblock Nuscenes-qa: A multi-modal visual question answering benchmark for autonomous driving scenario, 2024.

\bibitem[Qiao et~al.(2025)Qiao, Duan, Fang, Yang, Chen, Zhang, Wang, Lin, and Chen]{qiao2025prism}
Yuxuan Qiao, Haodong Duan, Xinyu Fang, Junming Yang, Lin Chen, Songyang Zhang, Jiaqi Wang, Dahua Lin, and Kai Chen.
\newblock Prism: A framework for decoupling and assessing the capabilities of vlms.
\newblock \emph{Advances in Neural Information Processing Systems}, 37:\penalty0 111863--111898, 2025.

\bibitem[Schwenk et~al.(2022)Schwenk, Khandelwal, Clark, Marino, and Mottaghi]{aokvqa2022}
Dustin Schwenk, Apoorv Khandelwal, Christopher Clark, Kenneth Marino, and Roozbeh Mottaghi.
\newblock A-okvqa: A benchmark for visual question answering using world knowledge.
\newblock In \emph{European Conference on Computer Vision (ECCV)}, 2022.

\bibitem[Shao et~al.(2024)Shao, Qian, Xiao, Song, Zong, Wang, Liu, and Li]{shao2024visual}
Hao Shao, Shengju Qian, Han Xiao, Guanglu Song, Zhuofan Zong, Letian Wang, Yu Liu, and Hongsheng Li.
\newblock Visual cot: Advancing multi-modal language models with a comprehensive dataset and benchmark for chain-of-thought reasoning.
\newblock \emph{arXiv preprint arXiv:2403.16999}, 2024.

\bibitem[Sima et~al.(2024)Sima, Renz, Chitta, Chen, Zhang, Xie, Luo, Geiger, and Li]{sima2023drivelm}
Chonghao Sima, Katrin Renz, Kashyap Chitta, Li Chen, Hanxue Zhang, Chengen Xie, Ping Luo, Andreas Geiger, and Hongyang Li.
\newblock Drivelm: Driving with graph visual question answering.
\newblock In \emph{European Conference on Computer Vision (ECCV)}, 2024.

\bibitem[Singh et~al.(2019)Singh, Natarjan, Shah, Jiang, Chen, Parikh, and Rohrbach]{singh2019towards}
Amanpreet Singh, Vivek Natarjan, Meet Shah, Yu Jiang, Xinlei Chen, Devi Parikh, and Marcus Rohrbach.
\newblock Towards vqa models that can read.
\newblock In \emph{Proceedings of the IEEE Conference on Computer Vision and Pattern Recognition (CVPR)}, 2019.

\bibitem[Sprague et~al.(2024)Sprague, Yin, Rodriguez, Jiang, Wadhwa, Singhal, Zhao, Ye, Mahowald, and Durrett]{sprague2024cot}
Zayne Sprague, Fangcong Yin, Juan~Diego Rodriguez, Dongwei Jiang, Manya Wadhwa, Prasann Singhal, Xinyu Zhao, Xi Ye, Kyle Mahowald, and Greg Durrett.
\newblock To cot or not to cot? chain-of-thought helps mainly on math and symbolic reasoning.
\newblock \emph{arXiv preprint arXiv:2409.12183}, 2024.

\bibitem[Sur{\'\i}s et~al.(2023)Sur{\'\i}s, Menon, and Vondrick]{suris2023vipergpt}
D{\'\i}dac Sur{\'\i}s, Sachit Menon, and Carl Vondrick.
\newblock Vipergpt: Visual inference via python execution for reasoning.
\newblock In \emph{Proceedings of the IEEE/CVF International Conference on Computer Vision}, pages 11888--11898, 2023.

\bibitem[Touvron et~al.(2023)Touvron, Lavril, Izacard, Martinet, Lachaux, Lacroix, Rozi{\`e}re, Goyal, Hambro, Azhar, et~al.]{touvron2023llama}
Hugo Touvron, Thibaut Lavril, Gautier Izacard, Xavier Martinet, Marie-Anne Lachaux, Timoth{\'e}e Lacroix, Baptiste Rozi{\`e}re, Naman Goyal, Eric Hambro, Faisal Azhar, et~al.
\newblock Llama: Open and efficient foundation language models.
\newblock \emph{arXiv preprint arXiv:2302.13971}, 2023.

\bibitem[Wang et~al.(2024{\natexlab{a}})Wang, Bai, Tan, Wang, Fan, Bai, Chen, Liu, Wang, Ge, Fan, Dang, Du, Ren, Men, Liu, Zhou, Zhou, and Lin]{Qwen2VL}
Peng Wang, Shuai Bai, Sinan Tan, Shijie Wang, Zhihao Fan, Jinze Bai, Keqin Chen, Xuejing Liu, Jialin Wang, Wenbin Ge, Yang Fan, Kai Dang, Mengfei Du, Xuancheng Ren, Rui Men, Dayiheng Liu, Chang Zhou, Jingren Zhou, and Junyang Lin.
\newblock Qwen2-vl: Enhancing vision-language model's perception of the world at any resolution.
\newblock \emph{arXiv preprint arXiv:2409.12191}, 2024{\natexlab{a}}.

\bibitem[Wang et~al.(2024{\natexlab{b}})Wang, Bai, Tan, Wang, Fan, Bai, Chen, Liu, Wang, Ge, et~al.]{wang2024qwen2}
Peng Wang, Shuai Bai, Sinan Tan, Shijie Wang, Zhihao Fan, Jinze Bai, Keqin Chen, Xuejing Liu, Jialin Wang, Wenbin Ge, et~al.
\newblock Qwen2-vl: Enhancing vision-language model's perception of the world at any resolution.
\newblock \emph{arXiv preprint arXiv:2409.12191}, 2024{\natexlab{b}}.

\bibitem[Wang et~al.(2024{\natexlab{c}})Wang, Yu, Jiang, Lan, Shi, Chang, Kautz, Li, and Alvarez]{wang2024omnidriveholisticllmagentframework}
Shihao Wang, Zhiding Yu, Xiaohui Jiang, Shiyi Lan, Min Shi, Nadine Chang, Jan Kautz, Ying Li, and Jose~M. Alvarez.
\newblock Omnidrive: A holistic llm-agent framework for autonomous driving with 3d perception, reasoning and planning, 2024{\natexlab{c}}.

\bibitem[Wang et~al.(2023)Wang, Maalouf, Xiao, Ban, Amini, Rosman, Karaman, and Rus]{wang2023driveanywheregeneralizableendtoend}
Tsun-Hsuan Wang, Alaa Maalouf, Wei Xiao, Yutong Ban, Alexander Amini, Guy Rosman, Sertac Karaman, and Daniela Rus.
\newblock Drive anywhere: Generalizable end-to-end autonomous driving with multi-modal foundation models, 2023.

\bibitem[Wei et~al.(2022)Wei, Wang, Schuurmans, Bosma, ichter, Xia, Chi, Le, and Zhou]{CoT2022}
Jason Wei, Xuezhi Wang, Dale Schuurmans, Maarten Bosma, brian ichter, Fei Xia, Ed Chi, Quoc~V Le, and Denny Zhou.
\newblock Chain-of-thought prompting elicits reasoning in large language models.
\newblock In \emph{Advances in Neural Information Processing Systems (NeurIPS)}, 2022.

\bibitem[Xu et~al.(2024)Xu, Zhang, Xie, Zhao, Guo, Wong, Li, and Zhao]{xu2024drivegpt4}
Zhenhua Xu, Yujia Zhang, Enze Xie, Zhen Zhao, Yong Guo, Kwan-Yee~K Wong, Zhenguo Li, and Hengshuang Zhao.
\newblock Drivegpt4: Interpretable end-to-end autonomous driving via large language model.
\newblock \emph{IEEE Robotics and Automation Letters}, 2024.

\bibitem[Yan et~al.(2024)Yan, Yang, Wu, Zhu, Yang, Li, Lin, Wang, McAuley, Gao, et~al.]{yan2024list}
An Yan, Zhengyuan Yang, Junda Wu, Wanrong Zhu, Jianwei Yang, Linjie Li, Kevin Lin, Jianfeng Wang, Julian McAuley, Jianfeng Gao, et~al.
\newblock List items one by one: A new data source and learning paradigm for multimodal llms.
\newblock \emph{arXiv preprint arXiv:2404.16375}, 2024.

\bibitem[Yue et~al.(2024)Yue, Ni, Zhang, Zheng, Liu, Zhang, Stevens, Jiang, Ren, Sun, Wei, Yu, Yuan, Sun, Yin, Zheng, Yang, Liu, Huang, Sun, Su, and Chen]{yue2023mmmu}
Xiang Yue, Yuansheng Ni, Kai Zhang, Tianyu Zheng, Ruoqi Liu, Ge Zhang, Samuel Stevens, Dongfu Jiang, Weiming Ren, Yuxuan Sun, Cong Wei, Botao Yu, Ruibin Yuan, Renliang Sun, Ming Yin, Boyuan Zheng, Zhenzhu Yang, Yibo Liu, Wenhao Huang, Huan Sun, Yu Su, and Wenhu Chen.
\newblock Mmmu: A massive multi-discipline multimodal understanding and reasoning benchmark for expert agi.
\newblock In \emph{Proceedings of the IEEE Conference on Computer Vision and Pattern Recognition (CVPR)}, 2024.

\bibitem[Zhai et~al.(2023)Zhai, Mustafa, Kolesnikov, and Beyer]{zhai2023sigmoid}
Xiaohua Zhai, Basil Mustafa, Alexander Kolesnikov, and Lucas Beyer.
\newblock Sigmoid loss for language image pre-training.
\newblock In \emph{Proceedings of the IEEE/CVF International Conference on Computer Vision}, pages 11975--11986, 2023.

\bibitem[Zhang et~al.(2024{\natexlab{a}})Zhang, Zhang, Li, hai zhao, Karypis, and Smola]{zhang2024multimodal}
Zhuosheng Zhang, Aston Zhang, Mu Li, hai zhao, George Karypis, and Alex Smola.
\newblock Multimodal chain-of-thought reasoning in language models.
\newblock \emph{Transactions on Machine Learning Research (TMLR)}, 2024{\natexlab{a}}.

\bibitem[Zhang et~al.(2024{\natexlab{b}})Zhang, Zhang, Li, Karypis, Smola, et~al.]{zhangmultimodal}
Zhuosheng Zhang, Aston Zhang, Mu Li, George Karypis, Alex Smola, et~al.
\newblock Multimodal chain-of-thought reasoning in language models.
\newblock \emph{Transactions on Machine Learning Research}, 2024{\natexlab{b}}.

\bibitem[Zheng et~al.(2023{\natexlab{a}})Zheng, Yang, Tang, Zhou, and Yang]{zheng2023ddCoT}
Ge Zheng, Bin Yang, Jiajin Tang, Hong-Yu Zhou, and Sibei Yang.
\newblock Ddcot: Duty-distinct chain-of-thought prompting for multimodal reasoning in language models.
\newblock \emph{Advances in Neural Information Processing Systems}, 36:\penalty0 5168--5191, 2023{\natexlab{a}}.

\bibitem[Zheng et~al.(2023{\natexlab{b}})Zheng, Chiang, Sheng, Zhuang, Wu, Zhuang, Lin, Li, Li, Xing, et~al.]{zheng2023judging}
Lianmin Zheng, Wei-Lin Chiang, Ying Sheng, Siyuan Zhuang, Zhanghao Wu, Yonghao Zhuang, Zi Lin, Zhuohan Li, Dacheng Li, Eric Xing, et~al.
\newblock Judging llm-as-a-judge with mt-bench and chatbot arena.
\newblock \emph{Advances in Neural Information Processing Systems}, 36:\penalty0 46595--46623, 2023{\natexlab{b}}.

\end{thebibliography}
}
\clearpage
\setcounter{page}{1}
\maketitlesupplementary

\appendix

This supplementary material contains the following sections:
\begin{itemize}
    \item A description of our pipeline to generate the interleaved explanations in our dataset (\cref{appx_subsec:interleaved_explanations});
    \item A description of our pipeline to annotate relevant entities (\cref{appx_subsec:auto_extract_pipeline});
    \item Examples showcasing representative samples and interleaved explanations (\cref{app_dataset_examples});
    \item Detail of the prompts used for training our models (\cref{appx_subsec:prompts});
    \item Implementation details and hyperparameter configurations for model training (\cref{appx_subsec:implementation});
    \item LLM-as-a-judge prompt for evaluation of reasoning correctness (\cref{appx_subsec:reasoning_eval});
    \item Examples comparing various model outputs, showcasing the strengths and limits of our methods (\cref{appx_subsec:outputs});
    \item A table comparing \textsc{DrivingVQA} with visual reasoning and autonomous driving datasets from the literature (\cref{appx_sec:benchmarks_comparison}).
\end{itemize}

\section{Dataset Specifications}
\label{appx_sec:details_dataset}

This section details the methodologies employed for the generation of interleaved explanations and the annotations of relevant entities.

\subsection{Pipeline for Generating Interleaved Explanations}
\label{appx_subsec:interleaved_explanations}
As described in \cref{subsec:relevant_entities}, for each visual question in our dataset, human experts use the explanations to identify and localize the key entities in the image that are required for answering the question. This leads to an average of 1.5 entities per image, and up to 5 entities. Then, as explained in \cref{subsec:interleave}, we use GPT-4o to match this list of relevant entities back with the explanation to obtain an interleaved explanation. In practice, we want to interleave each key entity -- whether it is the entity label, its bounding box coordinates, or the corresponding image tokens -- inside the explanation. In the rest of the section, we represent the interleaved explanation with bounding box coordinates appended next to the name of the entity referred to in the explanation.

To generate the interleaved explanations, we employ a strategy that combines few-shot prompting and cleaning heuristics. 

\paragraph{Initial interleaved explanation generation.} First, we feed GPT4o with each sample: the image, question, list of options, explanation, and the list of manually annotated entities along with bounding box coordinates. Following two demonstrations and a strict set of instructions, it is asked to interleave the bounding boxes of the entities inside the explanation, with minimal change to the explanation. Here, we provide the full instructions and the two hand-crafted demonstrations fed to the model.

\begin{tcolorbox}[colback=lightgray, colframe=darkblue, title=System prompt]
You are an expert at driving theory. You are tasked with helping a student answer questions about driving scenes.
\end{tcolorbox}

\begin{tcolorbox}[colback=lightgray, colframe=darkblue, title=Instructions]
    You are given an image of the driving theory test, a question about this image, a list of options, and the reasoning leading to the correct answer. 
    I also give you a list of entities along with bounding box coordinates, that are relevant to answering the question.

I need you to:
\begin{enumerate}
    \item Convert the original reasoning into a clear, step-by-step reasoning that makes use of all entities in the list.
    \item If the original reasoning does not mention one of the entities at all, add a short sentence referencing that entity and link it with the reasoning steps.
    \item Refer to each relevant entity in the format: **entity\_name** [x1, x2, y1, y2]. Replicate bounding box coordinates exactly as provided in the list.
    \item Keep all other wording as close to the original reasoning as possible.
    \item The entities should always be mentioned at the beginning of the sentences.
\end{enumerate}
\end{tcolorbox}
\vfill\null

\begin{tcolorbox}[colback=lightgray, colframe=darkblue, title=Demonstration 1]
\includegraphics[width=\linewidth]{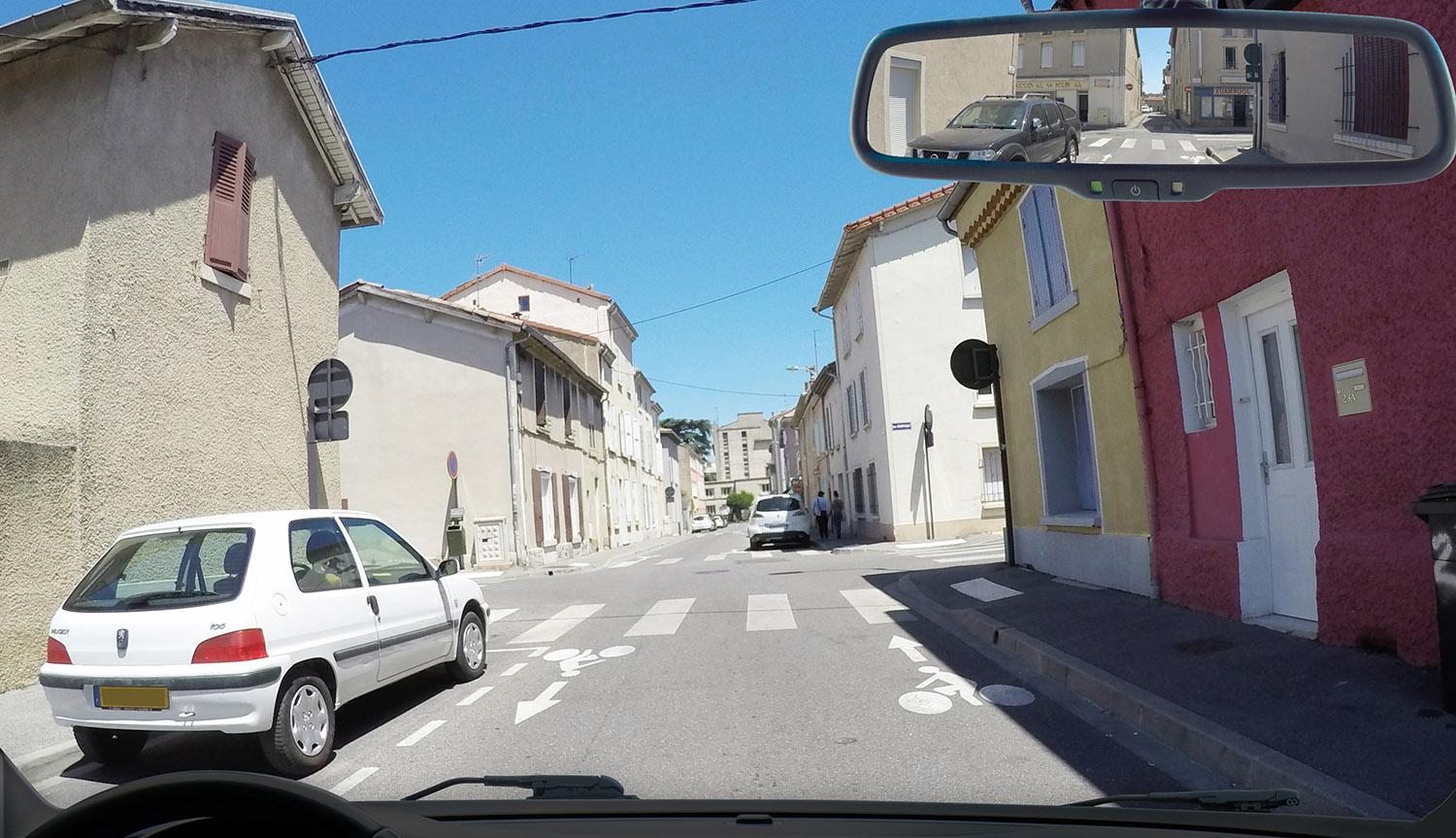}

\textbf{Question:} ["I am turning right at the next intersection. My attention is focused on: pedestrians", "I am turning right at the next intersection. My attention is focused on: cyclists"]

\textbf{Options:} \{A: Yes, B: No, C: Yes, D: No\}

\textbf{Answers:} A, C

\textbf{Reasoning:} Before turning right, I will approach a crosswalk. My attention is therefore on the arrival of pedestrians. As I am alongside a bike lane, I will check my right blind spot before turning my wheels to ensure that no cyclist is next to me.

\textbf{Entities:} 
\begin{itemize}
\item \textbf{pedestrian crossing}: [457.51,
     595.61,
     499.56,
     62.67]
\item \textbf{bike lane marking}: [893.4, 597.64, 196.2, 202.08]
   \end{itemize}
\end{tcolorbox}

\begin{tcolorbox}[colback=lightgray, colframe=darkred, title=Interleaved Reasoning]
Before turning right at the next intersection, I will approach a \textbf{\textcolor{darkblue}{pedestrian crossing}} [359.5, 337.11, 259.48, 22.09]. 
My attention is therefore on the arrival of pedestrians. As I am alongside a \textbf{\textcolor{darkblue}{bike lane marking}} [893.4, 597.64, 196.2, 202.08], 
I will check my right blind spot before turning my wheels to ensure that no cyclist is next to me.
\end{tcolorbox}
\vfill\null

\begin{tcolorbox}[colback=lightgray, colframe=darkblue, title=Demonstration 2]
\includegraphics[width=\linewidth]{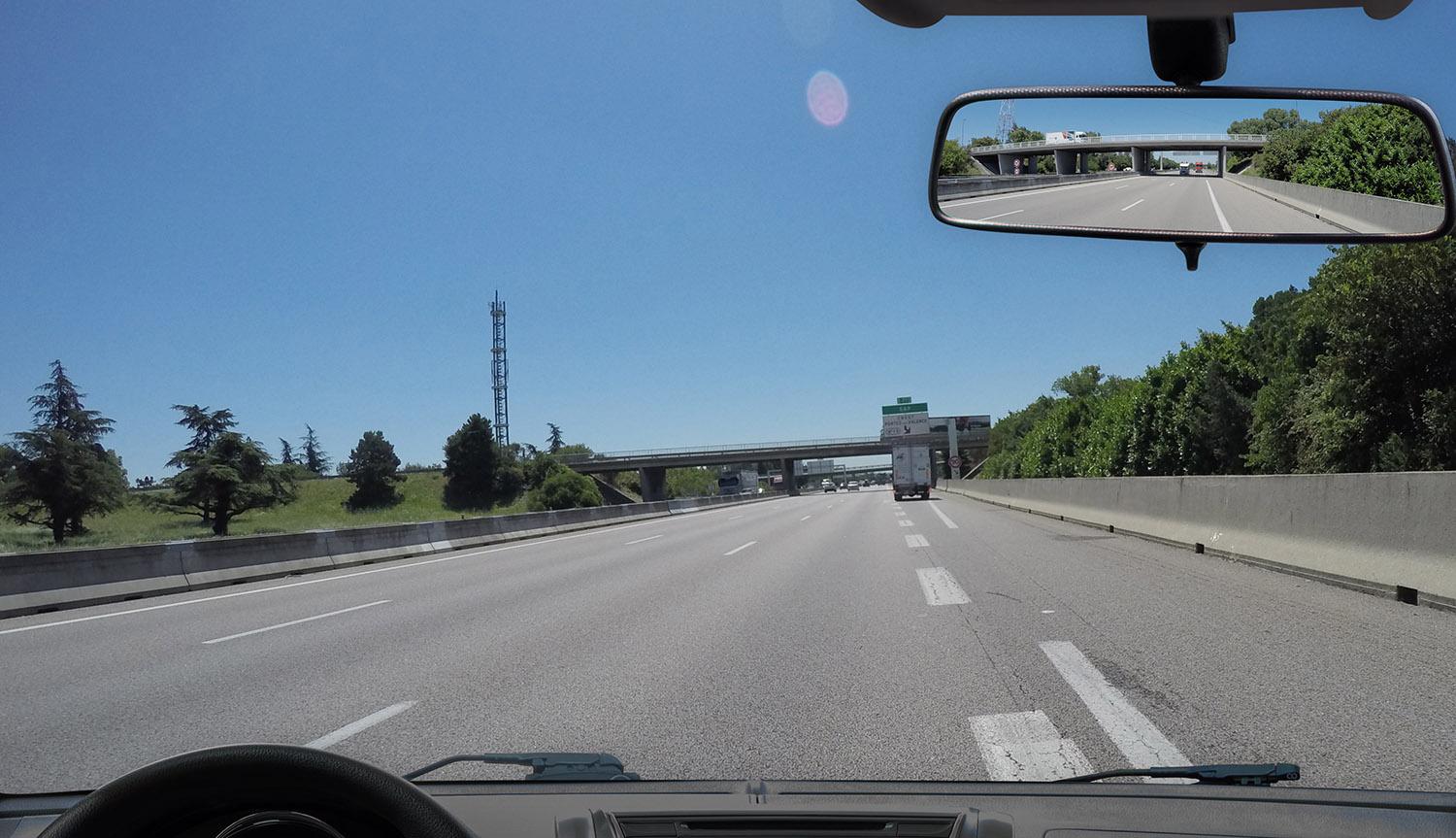}

\textbf{Question:} By taking this exit

\textbf{Options:} \{A: I can slow down right now, B: I must wait to brake\}

\textbf{Answers:} A

\textbf{Reasoning:} A deceleration lane allows me to exit without disrupting the flow of traffic behind me. The vehicle behind is far away, so I can start slowing down right now.

\textbf{Entities:} 
\begin{itemize}
    \item \textbf{deceleration lane:} [933.62, 522.54, 537.35, 249.17]
    \item \textbf{exit sign:} [904.6, 413.47, 57.03, 36.02]
    \item \textbf{vehicle:} [1206.9, 156.54, 44.8, 28.56]
    \item \textbf{rear-view mirror:} [952.52, 82.74, 544.27, 181.17]
\end{itemize}
\end{tcolorbox}

\begin{tcolorbox}[colback=lightgray, colframe=darkred, title=Interleaved Reasoning]
A \textbf{\textcolor{darkblue}{deceleration lane}} [933.62, 522.54, 537.35, 249.17] allows me to exit without disrupting the flow of traffic behind me. 
An \textbf{\textcolor{darkblue}{exit sign}} [904.6, 413.47, 57.03, 36.02] indicates the upcoming exit. The \textbf{\textcolor{darkblue}{vehicle}} [1206.9, 156.54, 44.8, 28.56] behind is far away, as I can see in the \textbf{\textcolor{darkblue}{rear-view mirror}} [952.52, 82.74, 544.27, 181.17]. 
So I can start slowing down right now.
\end{tcolorbox}

\vfill\eject

\paragraph{Explanations filtering and cleaning.} 

We clean the generated interleaved explanations using regular expressions and heuristics.
\begin{itemize}
\item We match the bounding box coordinates in the generated interleaved explanation with the ones in the list of entities provided, correcting minor variations due to the model failing to exactly replicate the set of coordinates.
\item We remove any hallucinated set of coordinates, that is absent from the provided list of entities.
    \item When an annotated bounding box was used twice in the interleaved explanation with different entity labels; if we can automatically identify the correct entity label, we remove the duplicated bounding box coordinates. Otherwise, we keep only the first occurrence of the set of coordinates.
\end{itemize}

\vfill\null

\begin{table*}[htbp]
    \centering
    \begin{tabularx}{\textwidth}{>{\bfseries}l X}
        \toprule
        Category & \textbf{Entities} \\
        \midrule
        \textbf{Road Signs} & speed limit sign, end of speed limit sign, yield sign, directional sign, stop sign, intersection sign, mandatory right turn sign, mandatory left turn sign, mandatory straight ahead sign, no entry sign, no right turn sign, no left turn sign, no u-turn sign, no overtaking sign, end of overtaking prohibition sign, danger sign, priority sign, exit sign, dead end road sign, merge lane sign, level crossing sign, two-way traffic sign, emergency phone sign, handicapped accessible sign, parking prohibition sign, end of restrictions sign, dimension restriction sign, road narrowing sign, one-way street sign, construction detour sign, pedestrian crossing sign, pedestrian underpass sign, school crossing sign, town entry sign, town exit sign, direction sign, wild animal crossing sign, construction sign, toll road sign, weather-related sign, camping zone sign, chevron alignment marker \\
        \addlinespace
        \textbf{Road Markings} & solid line, dashed line, pedestrian crossing, directional arrow, merge lane arrow, bike lane marking, stop line marking, loading zone line marking, traffic cones, temporary barrier \\
        \addlinespace
        \textbf{Road Features} & speed bump, roundabout, tunnel, bridge, construction zone, accident, emergency phone, toll booth, parking lane, bus lane, bus stop area, bicycle lane, emergency lane, entry lane, exit lane \\
        \addlinespace
        \textbf{Vehicles} & car, truck, motorcycle, bus, cyclist, van, motorhome, agricultural vehicle, public service vehicle, emergency vehicle \\
        \addlinespace
        \textbf{People and Animals} & pedestrian, police officer, construction worker, horse rider, animal \\
        \addlinespace
        \textbf{Vehicle Parts} & rear-view mirror, side-view mirror, turn signals, brake lights \\
        \bottomrule
    \end{tabularx}
    \caption{Categorized taxonomy of relevant entities for driving scenarios.}
    \label{appx_tab:entity_taxonomy}
\end{table*}

\begin{figure*}
    \begin{tcolorbox}[
        colback=gray!10,    
        colframe=black,     
        boxrule=0.5mm,      
        width=\textwidth,   
        arc=5mm,            
        auto outer arc,     
        left=4mm,           
        right=4mm,          
        top=2mm,            
        bottom=2mm          
    ]
            You are a driving theory expert, and your role is to extract entities from a driving scenario. These entities will be passed to an object detector for recognition. \\
        
        All the possible entities are:
        \begin{itemize}
            \item road signs: (See listed entities in \cref{appx_tab:entity_taxonomy})
            \item road markings: (See listed entities in \cref{appx_tab:entity_taxonomy})
            \item road features: (See listed entities in \cref{appx_tab:entity_taxonomy})
            \item vehicles: (See listed entities in \cref{appx_tab:entity_taxonomy})
            \item people and animals: (See listed entities in \cref{appx_tab:entity_taxonomy})
            \item vehicle parts: (See listed entities in \cref{appx_tab:entity_taxonomy})
        \end{itemize} 
        
        \vspace{0.3cm}
        \textbf{Instructions} \\
        Extract all entities from the scenes that are relevant to the following explanation and return them as a list. The output format should be only a list of entities, such as \texttt{[cyclist, oncoming vehicle, solid line, pedestrian crossing]}.
        Prioritize visible signs, markings, and vehicles directly affecting the scenario. If present in the image, always include any rear-view mirror or side-view mirror. 
        For help, you can also refer to the questions, possible answers and true answer below, as well as the provided image attached. \\
        
        \textbf{Question(s)}: \texttt{<questions\_text>} \\
        \textbf{Possible answers}: \texttt{<answers\_text>} \\
        \textbf{Correct answer}: \texttt{<correct\_answer\_letters>} \\
        \textbf{Explanation}: \texttt{<explanation\_text>} \\
    
    \end{tcolorbox}
    \caption{Prompt for relevant entity extraction.}
    \label{appx_fig:entity_extraction_prompt}
\end{figure*}

\subsection{Pipeline for Annotating Relevant Entities}
\label{appx_subsec:auto_extract_pipeline}

To enrich the collected dataset with relevant entities for each sample, we developed a semi-automated entity extraction and localization pipeline that helped to accelerate the annotation process. This pipeline comprises three steps, detailed below.

\paragraph{Initial entity extraction.} 
We define a taxonomy of potential entities commonly encountered in driving scenarios, organized into six groups (see \cref{appx_tab:entity_taxonomy}). This taxonomy includes categories such as road signs, road markings, vehicles, people and other objects. Given this taxonomy, a multi-modal language model, such as GPT-4o-mini, is prompted with instructions to extract for each sample an initial list of entities relevant to the human-expert explanation. The prompt also integrates textual cues from questions, possible answers, and correct answer along with the associated image. The prompt, illustrated in \cref{appx_fig:entity_extraction_prompt}, is designed to guide the model to prioritize visible and contextually significant entities. The extracted entities are returned in a structured list format, e.g., [\texttt{cyclist, pedestrian crossing, solid line}].

\paragraph{Relevant entities localization.}
After generating the list of relevant entities, we use a pre-trained object detection model, such as GroundingDINO~\cite{liu2023grounding}, to localize these entities within the images. This step provides a bounding box for each detected entity. Detected entities undergo refinement such as grouping semantically similar labels under a unified category and filtering out irrelevant or erroneous detections that deviate from the predefined taxonomy.

\paragraph{Manual validation.}
The output of the automated pipeline is validated and refined by human annotators. Irrelevant entities are removed, inaccurate labels are corrected, and missing entities are added to ensure dataset consistency and accuracy.

\newpage

\section{Examples}
\label{app_dataset_examples}

This section presents four representative examples from \textsc{DrivingVQA} dataset. Each example includes an egocentric image, one or two questions, a set of 2 to 4 answer options, the correct answers, a list of entities critical for answering the questions, and a reasoning explanation interleaved with the relevant entities. We also provide the original non-interleaved reasoning to enable a direct comparison.

The first two examples showcase where different entities have the same label, making the matching more challenging. Note that these two examples have question pairs, the first one associated with answer choices A and B, the second one having answer choices C and D.


\begin{tcolorbox}[colback=lightgray, colframe=darkgreen, title=Augmented Dataset Example 1]
\newcommand{\imgwidth}{900}
\newcommand{\imgheight}{350}
\drawImageWithBoxes{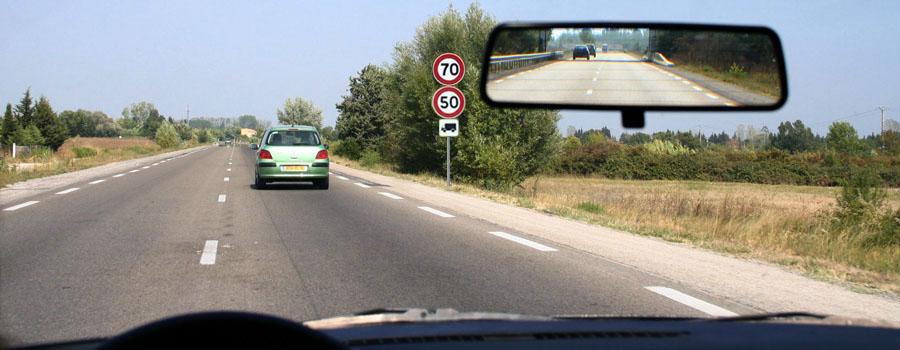}{900}{350}{
    \drawBoundingBox{429.54}{50.63}{38.06}{35.67}{\imgwidth}{\imgheight}{red}; 
    \drawBoundingBox{431.53}{86.7}{33.9}{49.8}{\imgwidth}{\imgheight}{blue}; 
}

\textbf{Question:} ["Can I drive at 50 km/h:", "70 km/h"]

\textbf{Options:} \{A: Yes, B: No, C: Yes, D: No\}

\textbf{Answers:} A, C

\textbf{Entities:} 
\begin{itemize}
    \item \textbf{speed limit sign}: [429.54, 50.63, 38.06, 35.67]
    \item \textbf{speed limit sign}: [431.53, 86.7, 33.9, 49.8]
\end{itemize}

\textbf{Reasoning}: The 2 speed limit signs are independent. The lower one limits the speed to 50 km/h only for vehicles designated for the transport of goods, as indicated by the category sign. The 70 km/h speed limit sign applies to all other categories of vehicles. I am not driving a goods transport vehicle, so I can drive at any speed not exceeding 70 km/h.

\textbf{Interleaved Reasoning:} The two speed limit signs are independent. The lower \textbf{\textcolor{blue}{speed limit sign}} [429.54, 50.63, 38.06, 35.67] limits the speed to 50 km/h only for vehicles designated for the transport of goods, as indicated by the category sign. The \textbf{\textcolor{red}{speed limit sign}} [431.53, 86.7, 33.9, 49.8] applies to all other categories of vehicles. I am not driving a goods transport vehicle, so I can drive at any speed not exceeding 70 km/h.

\end{tcolorbox}

\vfill\eject

\begin{tcolorbox}[colback=lightgray, colframe=darkgreen, title=Augmented Dataset Example 2]
\newcommand{\imgwidth}{1000}
\newcommand{\imgheight}{563}
\drawImageWithBoxes{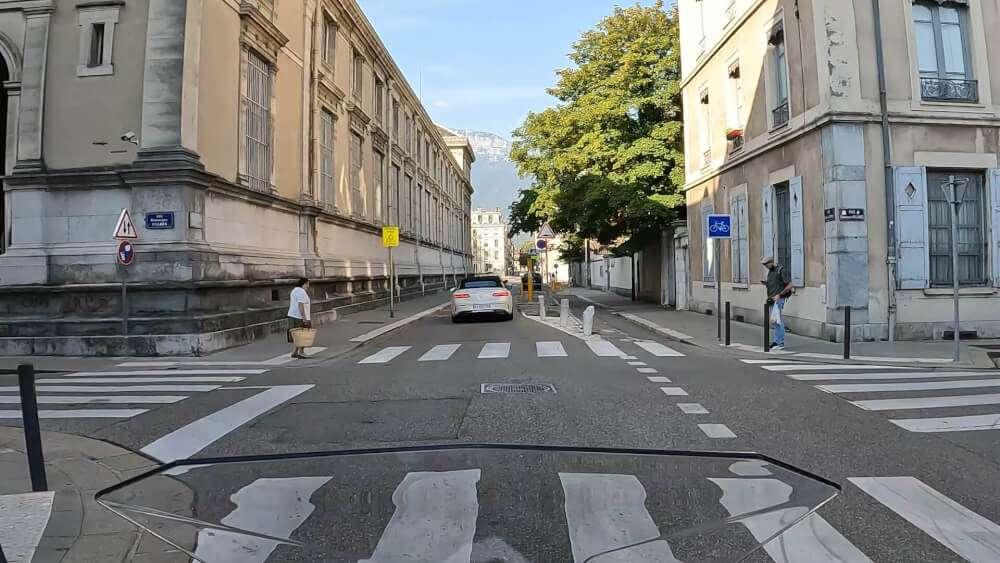}{1000}{563}{
    \drawBoundingBox{284.17}{274.3}{31.37}{84.98}{\imgwidth}{\imgheight}{red}; 
    \drawBoundingBox{359.5}{337.11}{259.48}{22.09}{\imgwidth}{\imgheight}{blue}; 
    \drawBoundingBox{757.46}{252.83}{37.6}{99.3}{\imgwidth}{\imgheight}{darkgreen}; 
}

\textbf{Question:} ["I monitor the pedestrian's intention:", "I prepare to stop:"]

\textbf{Options:} \{A: from the left, B: from the right, C: Yes, D: No\}

\textbf{Answers:} A, B, C

\textbf{Entities:} 
\begin{itemize}
    \item \textbf{pedestrian}: [284.17, 274.3, 31.37, 84.98]
    \item \textbf{pedestrian crossing}: [359.5, 337.11, 259.48, 22.09]
    \item \textbf{pedestrian}: [757.46, 252.83, 37.6, 99.3]
\end{itemize}

\textbf{Reasoning:}  The pedestrian on the left is very close to the crosswalk and is therefore preparing to cross. The pedestrian on the right is leaning over his phone and his attention is significantly diminished, so I am also monitoring him. I prepare to stop to let these two pedestrians cross.

\textbf{Interleaved Reasoning:} The \textbf{\textcolor{red}{pedestrian}} [284.17, 274.3, 31.37, 84.98] on the left is very close to the \textbf{\textcolor{blue}{pedestrian crossing}} [359.5, 337.11, 259.48, 22.09] and is therefore preparing to cross. The \textbf{\textcolor{darkgreen}{pedestrian}} [757.46, 252.83, 37.6, 99.3] on the right is leaning over his phone, and his attention is significantly diminished, so I am also monitoring him. I prepare to stop to let these two pedestrians cross.

\end{tcolorbox}

\vfill\eject

The following example shows a case where the explanation had to be modified, adding an extra sentence to include the relevant entity \textit{dashed line} that was manually annotated by the human experts but wasn't mentioned in the original explanation.

\begin{tcolorbox}[colback=lightgray, colframe=darkgreen, title=Augemented Dataset Example 3]
\newcommand{\imgwidth}{3824}
\newcommand{\imgheight}{1757}
\drawImageWithBoxes{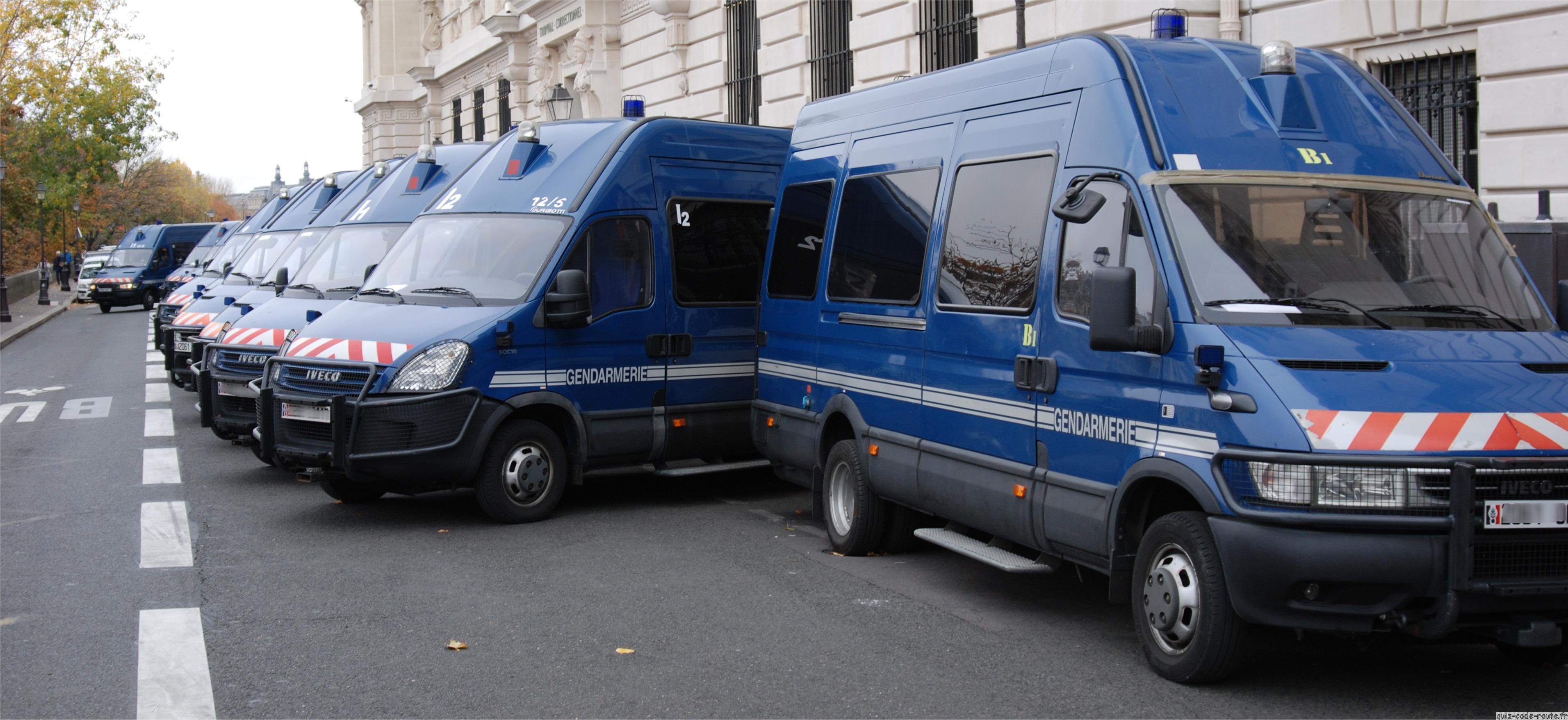}{3824}{1757}{
    \drawBoundingBox{372.0}{235.7}{1562.24}{1043.57}{\imgwidth}{\imgheight}{red}; 
    \drawBoundingBox{14.46}{755.5}{393.74}{996.79}{\imgwidth}{\imgheight}{blue}; 
    \drawBoundingBox{320.34}{747.66}{214.84}{998.63}{\imgwidth}{\imgheight}{darkgreen}; 
}

\textbf{Question:} Do the tradespeople run a risk if they park their van in the same way these police vans are parked?

\textbf{Options:} \{A: Yes, B: No\}

\textbf{Answer:} A

\textbf{Entities:} 
\begin{itemize}
    \item \textbf{police vehicle}: [372.0, 235.7, 1562.24, 1043.57]
    \item \textbf{bus lane}: [14.46, 755.5, 393.74, 996.79]
    \item \textbf{dashed line}: [320.34, 747.66, 214.84, 998.63]
\end{itemize}

\textbf{Reasoning:} These vans are not parked properly: they overflow so much from the spaces that road users have to move onto the bus lane, which is reserved for buses. We advise an average tradesperson against parking their van in this manner.

\textbf{Interleaved Reasoning:}  These \textbf{\textcolor{red}{police vehicles}} [372.0, 235.7, 1562.24, 1043.57] are not parked properly: they overflow so much from the spaces that road users have to move onto the \textbf{\textcolor{blue}{bus lane}} [14.46, 755.5, 393.74, 996.79], which is reserved for buses. The \textbf{\textcolor{darkgreen}{dashed line}} [320.34, 747.66, 214.84, 998.63] indicates the boundary of the lane. We advise an average tradesperson against parking their van in this manner.

\end{tcolorbox}

\vfill\eject

This final example shows a case with 5 annotated relevant entities, which is the maximal number that can be found in our dataset.  

\begin{tcolorbox}[colback=lightgray, colframe=darkgreen, title=Augmented Dataset Example 4]
\newcommand{\imgwidth}{1500}
\newcommand{\imgheight}{863}
\drawImageWithBoxes{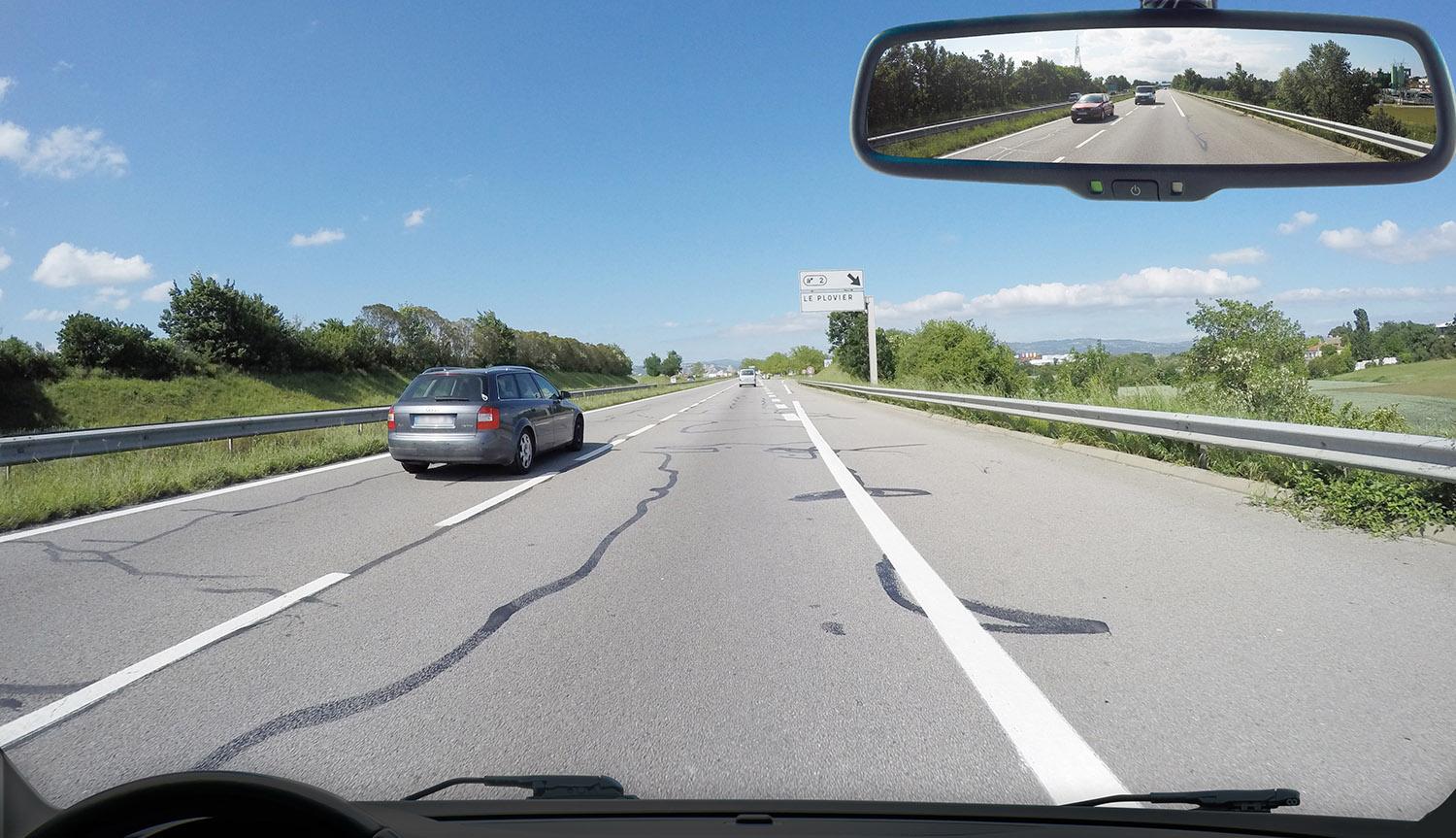}{1500}{863}{
    \drawBoundingBox{317.36}{423.1}{394.94}{188.64}{\imgwidth}{\imgheight}{red}; 
    \drawBoundingBox{819.58}{272.31}{75.71}{51.58}{\imgwidth}{\imgheight}{blue}; 
    \drawBoundingBox{1101.09}{77.8}{93.61}{49.79}{\imgwidth}{\imgheight}{darkgreen}; 
    \drawBoundingBox{873.99}{9.21}{623.74}{196.48}{\imgwidth}{\imgheight}{purple}; 
    \drawBoundingBox{394.01}{371.19}{211.56}{119.92}{\imgwidth}{\imgheight}{orange}; 
}

\textbf{Question:} ["The vehicle in front can still take the next exit", "I can follow this vehicle to overtake"]

\textbf{Options:} \{A: Yes, B: No, C: Yes, D: No\}

\textbf{Answers:} B, D

\textbf{Entities:} 
\begin{itemize}
    \item \textbf{dashed line}: [317.36, 423.1, 394.94, 188.64]
    \item \textbf{exit sign}: [819.58, 272.31, 75.71, 51.58]
    \item \textbf{car}: [1101.09, 77.8, 93.61, 49.79]
    \item \textbf{rear-view mirror}: [873.99, 9.21, 623.74, 196.48]
    \item \textbf{car}: [394.01, 371.19, 211.56, 119.92]
\end{itemize}

\textbf{Reasoning:} The marking in the middle of the two lanes of traffic only prohibits vehicles in the left lane from merging to take the exit. However, vehicles are approaching from behind, so I must give up overtaking.

\textbf{Interleaved Reasoning:} The \textbf{\textcolor{red}{dashed line}} [317.36, 423.1, 394.94, 188.64] in the middle of the two lanes of traffic only prohibits \textbf{\textcolor{orange}{vehicles in the left lane}} [394.01, 371.19, 211.56, 119.92] from merging to take the exit indicated by the \textbf{\textcolor{blue}{exit sign}} [819.58, 272.31, 75.71, 51.58]. However, vehicles [\textbf{\textcolor{darkgreen}{1101.09, 77.8, 93.61, 49.79}}] are approaching from behind, as seen in the \textbf{\textcolor{purple}{rear-view mirror}} [873.99, 9.21, 623.74, 196.48], so I must give up overtaking.

\end{tcolorbox}


\newpage

\section{Experiments Details}
\label{appx_sec:experiments}
\subsection{Format-Specific Prompt Templates}
\label{appx_subsec:prompts}
We begin by providing the detailed prompting format for each variant (QPR-EA, QPRB-EA, QPRV-EA) to analyze which signal is the most relevant to enhance visual-chain-of-thought reasoning. Note that the bounding boxes are represented in the format $(x, y, w, h)$, where $(x, y)$ specifies the coordinates of the top-left corner, and $w$ and $h$ denote the width and height of the bounding box. Similarly to \cite{chen2023shikra} we normalize the bounding boxes between $0$ and $1$ during training and generation. The following example illustrates the QPR-EA and the QPRB-EA prompting strategies:

\begin{tcolorbox}[colback=lightgray, colframe=black, title=QP\textcolor{cyan}{R}\textcolor{ForestGreen}{B}-EA]
\newcommand{\imgwidth}{1500}
\newcommand{\imgheight}{863}
\drawImageWithBoxes{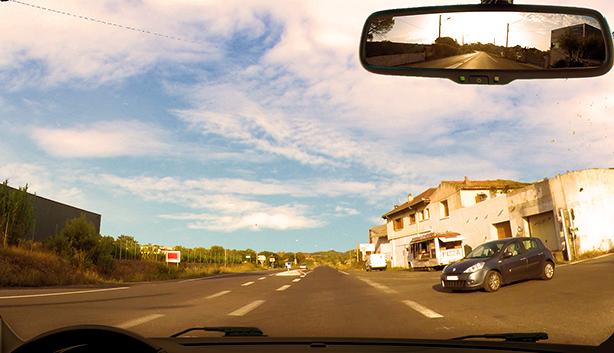}{1500}{863}

\textbf{Human:}\\ Unless explicitly stated otherwise, assume you are driving a car in France.\\
Select all correct answers to the following question from the available options. Detail your reasoning step by step based on road signs, markings, signals, and relevant driving rules. Then, provide the letters corresponding to your answer in the format: 'Answer(s): $<$letters$>$' \\
Question: As I approach this intersection, I position my foot above the brake
Options: \{A: Yes, B No.\} \\
The relevant entities for this problem are: \textcolor{cyan}{car} \textcolor{ForestGreen}{[0.715, 0.665, 0.908, 0.830]}. \\
\textbf{VLM:} \\
Reasoning: At the level of this intersection, there is a car on the right. I don’t know if the driver has seen me since he is not looking at me. Therefore, I position my foot in a safe position above the brake so that I can stop if the driver decides to proceed. \\
Answer(s): A.
\end{tcolorbox}

\vfill\eject

For QPRV-EA, we use a prompting strategy with visual tokens denoted $<$image$>$ that is the features of the relevant entity's bounding box from the SigLIP visual encoder. It corresponds to:

\begin{tcolorbox}[colback=lightgray, colframe=black, title=QP\textcolor{cyan}{R}\textcolor{Fuchsia}{V}-EA]
\newcommand{\imgwidth}{1500}
\newcommand{\imgheight}{863}
\drawImageWithBoxes{images/examples/0394.jpg}{1500}{863}

\textbf{Human:}\\ Unless explicitly stated otherwise, assume you are driving a car in France.\\
Select all correct answers to the following question from the available options. Detail your reasoning step by step based on road signs, markings, signals, and relevant driving rules. Then, provide the letters corresponding to your answer in the format: 'Answer(s): $<$letters$>$' \\
Question: As I approach this intersection, I position my foot above the brake: \\
Options: \{A: Yes, B No.\} \\
The relevant entities for this problem are: \textcolor{cyan}{car} \textcolor{Fuchsia}{$<$ image$>$}. \\
\includegraphics[width=0.2\textwidth]{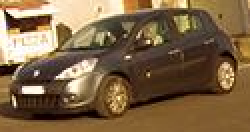} \\
\textbf{VLM:} \\
Reasoning: At the level of this intersection, there is a car on the right. I don’t know if the driver has seen me since he is not looking at me. Therefore, I position my foot in a safe position above the brake so that I can stop if the driver decides to proceed. \\
Answer(s): A.
\end{tcolorbox}

\vfill\eject

Then, we provide the prompts used for the two-step procedure outlined in \cref{subsec:learning_to_generate} to evaluate the ability of VLMs to predict relevant entities under different formats (name: QP-REA, localization: QP-RBEA, visual content: QP-RB-RV-EA). The following illustrates QR-REA and QP-RBEA:

\begin{tcolorbox}[colback=lightgray, colframe=black, title=QP-\textcolor{cyan}{R}\textcolor{ForestGreen}{B}EA]
\newcommand{\imgwidth}{1500}
\newcommand{\imgheight}{863}
\drawImageWithBoxes{images/examples/0394.jpg}{1500}{863}

\textbf{Human:}\\ 
Unless explicitly stated otherwise, assume you are driving a car in France. List all relevant entities from the scene that are necessary to answer the following question, such as road signs, markings, signals, or other vehicles in the image, along with their bounding boxes. Then, select all correct answers to the following question from the available options. Detail your reasoning step by step based on these entities and relevant driving rules. Provide the letters corresponding to your answer in the format: 'Answer(s): $<$letters$>$' \\
Question: As I approach this intersection, I position my foot above the brake: \\
Options: \{A: Yes, B No.\} \\
\textbf{VLM:} \\
The relevant entities for this problem are: \textcolor{cyan}{car} \textcolor{ForestGreen}{[0.715, 0.665, 0.908, 0.830]}. \\
Reasoning: At the level of this intersection, there is a car on the right. I don’t know if the driver has seen me since he is not looking at me. Therefore, I position my foot in a safe position above the brake so that I can stop if the driver decides to proceed. \\
Answer(s): A.
\end{tcolorbox}

\vfill\eject

Finally, the example below shows how to prompt the two-step procedure for QP-RB-RV-EA.

\begin{tcolorbox}[colback=lightgray, colframe=black, title=QP-\textcolor{cyan}{R}\textcolor{ForestGreen}{B}-\textcolor{cyan}{R}\textcolor{Fuchsia}{V}-EA]
\newcommand{\imgwidth}{1500}
\newcommand{\imgheight}{863}
\drawImageWithBoxes{images/examples/0394.jpg}{1500}{863}

\textbf{Human:}\\ Unless explicitly stated otherwise, assume you are driving a car in France.\\
List all relevant entities from the scene that are necessary to answer the following questions, such as road signs, markings, signals, or other vehicles in the image \\
Question: As I approach this intersection, I position my foot above the brake: \\
Options: \{A: Yes, B No.\} \\
\textbf{VLM:} \\
The relevant entities for this problem are: \textcolor{cyan}{car} \textcolor{ForestGreen}{[0.715, 0.665, 0.908, 0.830]}. \\
\textbf{Human:}\\ 
Their corresponding image patches are: \textcolor{cyan}{car} \textcolor{Fuchsia}{$<$image$>$}. 
\includegraphics[width=0.2\textwidth]{images/examples/0394_1.png}\\
Then, select all correct answers to the following questions from the available options. choose at least one answer per question. Detail your reasoning step by step based on these entities and relevant driving rules. Provide the letters corresponding to your answer in the format: 'Answer(s): $<$letters$>$'. \\
\textbf{VLM:} \\
Reasoning: At the level of this intersection, there is a car on the right. I don’t know if the driver has seen me since he is not looking at me. Therefore, I position my foot in a safe position above the brake so that I can stop if the driver decides to proceed. \\
Answer(s): A.
\end{tcolorbox}

\vfill\eject

\subsection{Implementation Details}
\label{appx_subsec:implementation}

We fine-tune various LLaVA-OneVision models using different prompt formats on the \textsc{DrivingVQA} train set. All models are initialized from the pre-trained weights obtained after the final OneVision training stage.

The fine-tuning experiments are conducted using 2 NVIDIA A100 GPUs, except for experiments involving visual patches ('V') where 4 NVIDIA A100 GPUs were required. We employ the following configuration and hyper-parameters:

\begin{table}[ht]
\centering
\setlength{\tabcolsep}{8pt} 
\renewcommand{\arraystretch}{1.2} 
 \resizebox{\linewidth}{!}{%
\begin{tabular}{ll}
\toprule
\textbf{Hyperparameter}       & \textbf{Value} \\ 
\midrule
\textbf{Epochs}               & $10$ \\
\textbf{Batch Size}           & $2$ \\
\textbf{Max Seq. Length}  & $32768$ \\
\midrule
\textbf{Vision Tower}         & SigLIP-SO400M-Patch14-384 \\
\textbf{Language Model}       & Qwen2-7B \\
\textbf{Projector}       & MLP2x\_GELU \\
\midrule
\textbf{Trainable Parts}      & $\phi_{\text{vision}}$, $\theta_{\text{proj}}$, $\theta_{\text{LLM}}$ \\
\textbf{Learning Rate} $\theta_{\text{proj}}$, $\theta_{\text{LLM}}$       & $1e^{-5}$ \\ 
\textbf{Learning Rate} $\phi_{\text{vision}}$        & $2e^{-6}$ \\ 
\textbf{Weight Decay}         & $0$ \\
\textbf{LR Scheduler} & cosine \\
\textbf{Warm-Up Ratio}         & $0.03$ \\
\bottomrule
\end{tabular}
}
\caption{Hyper-parameter configuration for fine-tuning LLaVA-OneVision on the \textsc{DrivingVQA} dataset}
\label{tab:hyperparameters_finetuning}
\end{table}

\subsection{Model reasoning evaluation}
\label{appx_subsec:reasoning_eval}
\begin{figure*}[h]
\begin{tcolorbox}[colback=lightgray, colframe=darkblue, title=Instructions]
    \textbf{You are a strict but fair driving-theory instructor.} You’re given:
    \begin{enumerate}
        \item A driving theory test question
        \item A list of possible answer options
        \item The official “correct reasoning”
        \item A student’s reasoning for the same question
    \end{enumerate}

    \bigskip
    \textbf{Your task:} \textit{Assess if the student’s reasoning matches the correct reasoning.}

    \bigskip
    \textbf{Step-by-Step Instructions:}
    \begin{enumerate}
        \item \textbf{Identify Student’s Arguments}
        \begin{itemize}
            \item List each key argument or step in the student's reasoning.
            \item For each argument, briefly state whether it is correct or not, given the provided correct reasoning.
        \end{itemize}
        \item \textbf{Check for Missing or Contradictory Points}
        \begin{itemize}
            \item Look at the official correct reasoning.
            \item List \textbf{important points or steps} from the correct reasoning that the student \textbf{omits} or \textbf{directly contradicts}.
            \item Minor omissions or differences in wording/style are acceptable.
        \end{itemize}
        \item \textbf{Decide on Overall Correctness}
        \begin{itemize}
            \item If the student’s reasoning is \textbf{mostly consistent} with the correct reasoning and has \textbf{no major factual errors}, then it is considered \textbf{correct}.
            \item If the student’s reasoning \textbf{contains significant logical or factual errors}, or \textbf{omits critical steps} from the correct reasoning, then mark it \textbf{incorrect}.
        \end{itemize}
    \end{enumerate}

    \bigskip
    \textbf{Important Note:} The student’s reasoning does not have to match the official reasoning exactly; it just needs to be conceptually equivalent and free of serious contradictions.

    \bigskip
    \textbf{Final Output Format:}
    \begin{itemize}
        \item Provide your step-by-step analysis.
        \item At the end, write: \textbf{Final Answer}: “1” if you judge the student’s reasoning is overall correct, “0” if it is overall incorrect.
    \end{itemize}
    \end{tcolorbox}
    \caption{Prompt for evaluation of model reasoning.}
    \label{appx_fig:reasoning_eval_prompt}
\end{figure*}

\subsection{Result Output Examples}
\label{appx_subsec:outputs}

In this section, we present test samples alongside selected model outputs to highlight their capabilities and their limitations. For each sample, we provide the question and answer choices based on the QP-EA prompt (blue box), followed by the reasoning and answers from two models, along with their prompt-specific formats where applicable (red boxes).

\begin{tcolorbox}[colback=lightgray, colframe=darkblue, title=Question Example 1]
\includegraphics[width=\linewidth]{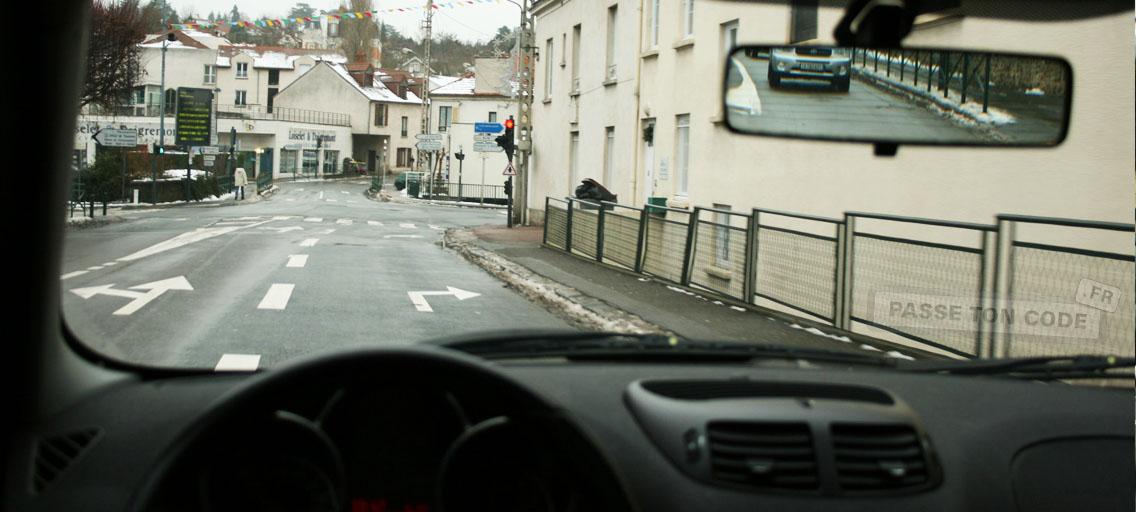}
\textbf{Human:}\\ Unless explicitly stated otherwise, assume you are driving a car in France.\\
Select all correct answers to the following question from the available options. Detail your reasoning step by step based on road signs, markings, signals, and relevant driving rules. Then, provide the letters corresponding to your answer in the format: 'Answer(s): $<$letters$>$' \\
Question: To go straight, can I stay in this lane? \\
Options: \{A: Yes, B No.\} \\
\end{tcolorbox}

\begin{tcolorbox}[colback=lightgray, colframe=darkred, title=Question Ouput 1: \textbf{QP-EA}]
\textbf{QP-EA:} \\
Reasoning: If I want to go straight, the road markings indicate that I am correctly positioned, as I can either continue straight or turn right. The left lane is the lane for turning left. Therefore, I will not need to change lanes.\\
Answer(s): A.
\end{tcolorbox}

\begin{tcolorbox}[colback=lightgray, colframe=darkred, title=Question Ouput 1: \textbf{QP\textcolor{cyan}{R}\textcolor{Fuchsia}{V}-EA}]
\textbf{Human:}\\
The relevant entities for this problem are: \textcolor{cyan}{directional arrow} 
\includegraphics[width=0.2\textwidth]{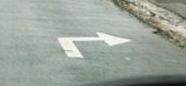}. \\

\textbf{QP\textcolor{cyan}{R}\textcolor{Fuchsia}{V}-EA:} \\
Reasoning: The lane I am in is intended for users wishing to turn right. This is indicated by the directional arrow on the road markings. Therefore, I cannot stay in this lane if my intention is to go straight.\\
Answer(s): B.
\end{tcolorbox}

In Example 1 above, the simple chain-of-thought model (QP-EA) acknowledges the existence of a directional arrow as road marking, but incorrectly hallucinates its direction, leading to the wrong answer. The ``Oracle'' model QPRV-EA, being fed with an image patch showing the arrow, is able to identify its direction and correctly answer the question.  The ground truth explanation is the following:\\
``\textit{The road markings indicate that the right lane can only be used for turning right. Therefore, I must position myself in the left lane to go straight or to the left.}''

\vfill\eject
\vfill\null

\begin{tcolorbox}[colback=lightgray, colframe=darkblue, title=Question Example 2]
\includegraphics[width=\linewidth]{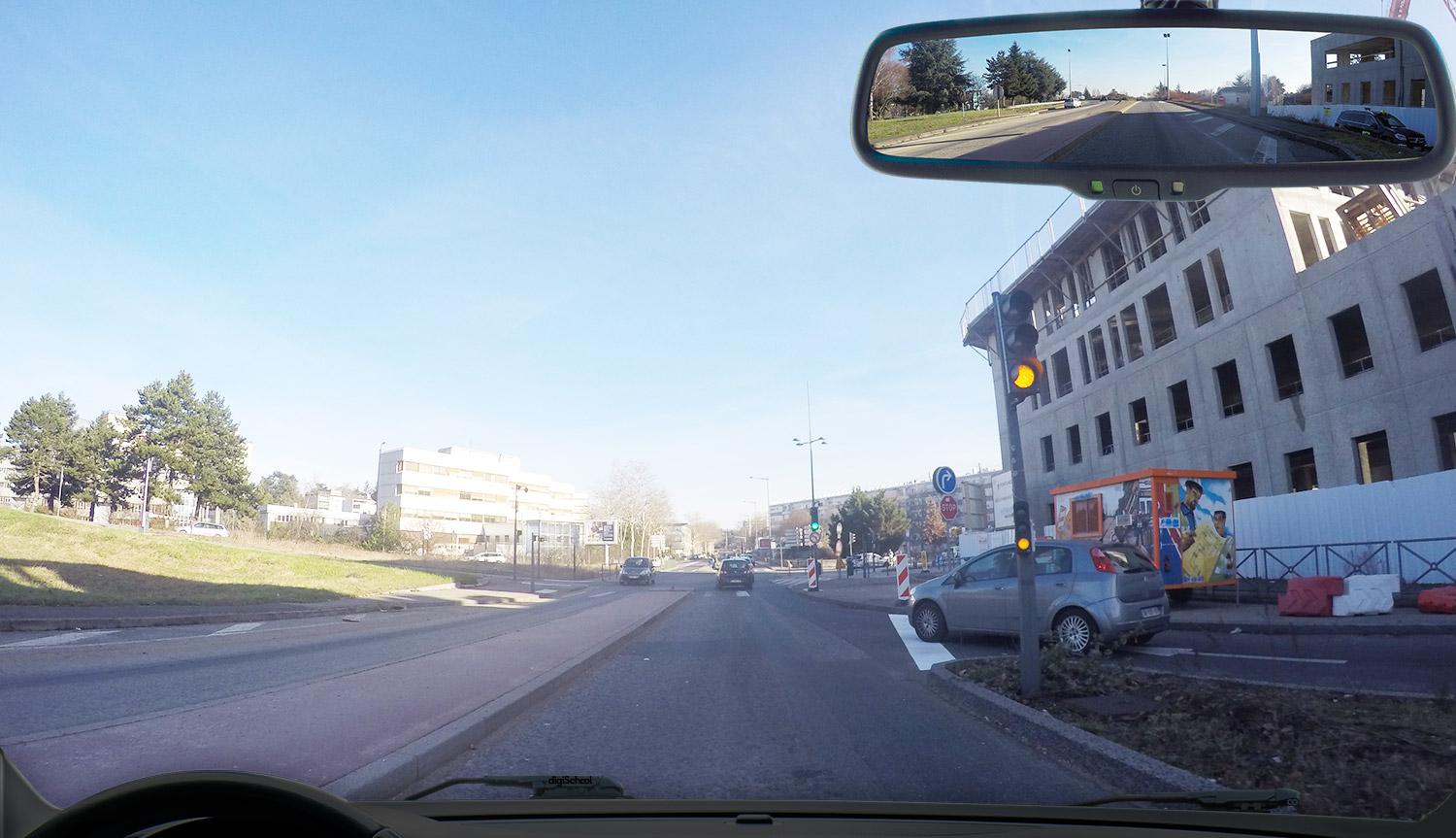}
\textbf{Human:}\\ Unless explicitly stated otherwise, assume you are driving a car in France.\\
Select all correct answers to the following question from the available options. Detail your reasoning step by step based on road signs, markings, signals, and relevant driving rules. Then, provide the letters corresponding to your answer in the format: 'Answer(s): $<$letters$>$' \\
Question: This light is flashing, I must yield to this car: \\
Options: \{A: Yes, B No.\} \\
\end{tcolorbox}

\begin{tcolorbox}[colback=lightgray, colframe=darkred, title=Question Output 1: \textbf{QP-\textcolor{cyan}{R}\textcolor{darkgreen}{B}EA}]
\textbf{QP-\textcolor{cyan}{R}\textcolor{darkgreen}{B}EA:} \\
The relevant entities for this problem are: car [0.617, 0.664, 0.754, 0.798], traffic light [0.652, 0.398, 0.679, 0.521].
Reasoning: This car is stopped at the traffic light, so it does not have the right of way. Therefore, I do not need to yield to it while waiting for the light to turn green.\\
Answer(s): B.
\end{tcolorbox}

\begin{tcolorbox}[colback=lightgray, colframe=darkred, title=Question Output 2: \textbf{QP\textcolor{cyan}{R}\textcolor{darkgreen}{B}-EA}]
\textbf{Human:}\\
The relevant entities for this problem are: traffic light [0.682, 0.344, 0.717, 0.662], vehicle [0.623, 0.640, 0.805, 0.784], stop line marking [0.616, 0.721, 0.655, 0.804], stop sign [0.640, 0.585, 0.661, 0.628], traffic light [0.555, 0.603, 0.564, 0.635].\\
\textbf{QP\textcolor{cyan}{R}\textcolor{darkgreen}{B}-EA:} \\
Reasoning: The traffic light is malfunctioning, and the vehicle that wants to turn left must stop at the stop line marking. It is not a priority, so I do not have to yield to it. The stop sign is also present.\\
Answer(s): B.
\end{tcolorbox}

In example 2, we compare providing the entities and bounding box coordinates as input (QPRB-EA), or letting the model detect them itself (QP-RBEA). Both models reach the same answer, but the second one's reasoning is wrong: it fails to identify a key entity (the stop line marking or the stop sign), leading to an incorrect reasoning even though the answer is correct. This highlights the importance of accurate entity detection in the reasoning process. The ground truth explanation is the following:\\
``\textit{This traffic light is flashing, and there is no sign under the light. However, I can see that the vehicle on the right has a stop line marking and a stop sign. There is also another traffic light visible. Therefore, I do not have to yield to the right.}''

\begin{tcolorbox}[colback=lightgray, colframe=darkblue, title=Question Example 3]
\includegraphics[width=\linewidth]{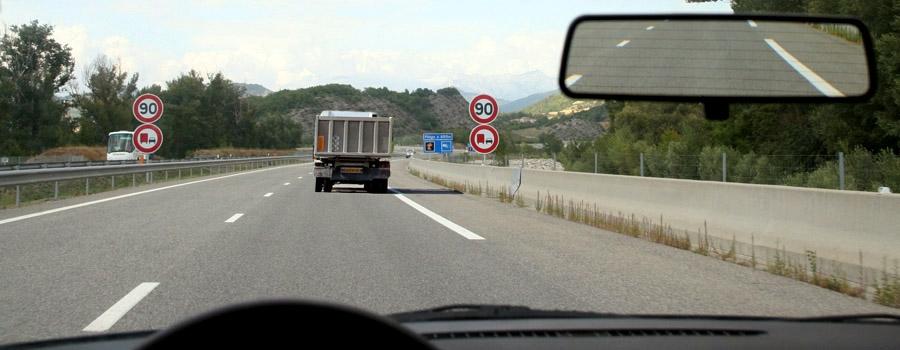}
\textbf{Human:}\\ Unless explicitly stated otherwise, assume you are driving a car in France.\\
Select all correct answers to the following question from the available options. Detail your reasoning step by step based on road signs, markings, signals, and relevant driving rules. Then, provide the letters corresponding to your answer in the format: 'Answer(s): $<$letters$>$' \\
Question: After these signs, will I be able to overtake this truck: \\
Options: \{A: Yes, B No.\} \\
\end{tcolorbox}

\begin{tcolorbox}[colback=lightgray, colframe=darkred, title=Question Ouput 1: \textbf{QP-EA}]
\textbf{QP-EA:} \\
Reasoning: The left lane is a storage lane that allows for left turns. It is strictly prohibited to use this type of lane for overtaking the truck. I therefore give up on making this maneuver.\\
Answer(s): B.
\end{tcolorbox}

\begin{tcolorbox}[colback=lightgray, colframe=darkred, title=Question Ouput 2: \textbf{QP-\textcolor{cyan}{R}\textcolor{darkgreen}{B}-\textcolor{cyan}{R}\textcolor{Fuchsia}{V}-EA}]

\textbf{QP-\textcolor{cyan}{R}\textcolor{darkgreen}{B}-\textcolor{cyan}{R}\textcolor{Fuchsia}{V}-EA:}
The relevant entities for this problem are: speed limit sign [0.508, 0.294, 0.547, 0.401], truck [0.367, 0.304, 0.461, 0.546].

\textbf{Human:}\\
The relevant entities for this problem are: \textcolor{cyan}{speed limit sign} \includegraphics[width=0.2\textwidth]{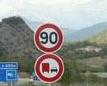}, \textcolor{cyan}{truck} \includegraphics[width=0.2\textwidth]{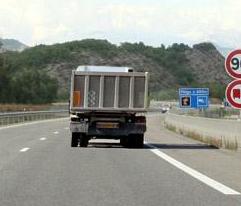}.\\

\textbf{QP-\textcolor{cyan}{R}\textcolor{darkgreen}{B}-\textcolor{cyan}{R}\textcolor{Fuchsia}{V}-EA:}\\
Reasoning: The prohibition signs must be read separately. The speed limit sign of 90 km/h applies to all road users. The second sign prohibits overtaking for vehicles transporting goods over 3.5 tons. Therefore, I will be able to overtake this truck.\\
Answer(s): A.
\end{tcolorbox}

Example 3 shows a case where despite an imperfect entity detection (``speed limit sign'' instead of ``no overtaking sign'', adding the image patches allows the model to correctly identify the important sign to answer the question. This is enabled by our cropping strategy that adds 70\% of the image patch to increase the context and account for imperfect entity detection. 
On the contrary, the baseline QP-EA hallucinates the presence of a storage lane for turning left and answers incorrectly.

The ground truth explanation is:
\textit{This no overtaking sign prohibits goods transport vehicles from overtaking. It does not apply to me. Furthermore, I am driving on a one-way road with 2 lanes, so I can use the left lane to overtake. It is clear, as confirmed by a quick glance in the rearview mirror. I can therefore proceed with my maneuver while ensuring that I respect the speed limit of 90 km/h.
}

\section{Benchmarks comparison}
\label{appx_sec:benchmarks_comparison}

\begin{table*}[t]
\centering
\renewcommand{\arraystretch}{1.0} 
\setlength{\tabcolsep}{6pt} 
\resizebox{\linewidth}{!}{%
\begin{tabular}{@{}lcccccrr@{}}
\toprule
\textbf{Dataset} & \textbf{Real} & \textbf{Explanations} & \textbf{BBoxes} & \textbf{Relevant entities} & \textbf{Interleaved expl.} & \textbf{\#Images} & \textbf{\#QAs} \\ 
\midrule
Causal-VidQA~\citep{li2022from} & \checkmark & human-annotated & \checkmark & \xmark & \xmark & 27K & 107K \\
ScienceQA~\cite{lu2022learn}  & $\sim$ & human-annotated  & \xmark & \xmark  & \xmark  & 10K & 21K \\
PuzzleQA~\citep{chia2024puzzlevqa}  & \xmark & LLM-generated    & \xmark & \xmark  & \xmark  & 2K & 2K  \\
GQA-CoT~\citep{shao2024visual}  & \checkmark & LLM-generated    & \checkmark  & single       & \xmark  & 88K & 88K \\
NuScenesQA~\citep{qian2024nuscenesqa} & \checkmark & \xmark & \checkmark & \xmark & \xmark & 34K & 460K \\
DriveLM-nuScenes~\citep{sima2023drivelm} & \checkmark & \xmark & \checkmark & \xmark & \xmark & 30K & 443K \\
BDD-X~\citep{kim2018textual} & \checkmark & human-annotated & \xmark & \xmark & \xmark & 7K & 26K \\
LingoQA~\citep{marcu2024lingoqa} & \checkmark & LLM-generated & \xmark & \xmark & \xmark & 28K & 420K \\
\midrule
\rowcolor{gray!10}
\textbf{\textsc{DrivingVQA}} (Ours) & \checkmark & human-annotated   & \checkmark  & multiple   & \checkmark   & 4K & 4K  \\ \bottomrule
\end{tabular}}
\caption{\textbf{Comparison of existing VQA datasets with explanations and/or for autonomous driving.} \emph{Real} designed real-word images datasets. \textsc{DrivingVQA} uniquely combines real-world images from driving scenarios with (1) human-annotated, interleaved explanations and (2) multiple relevant entity annotations.}
\label{tab:datasets}
\end{table*}

\end{document}